\documentclass[lettersize,journal]{IEEEtran}
\usepackage{amsmath,amsfonts}
\usepackage{algorithmic}
\usepackage{algorithm}
\usepackage{array}
\usepackage[caption=false,font=normalsize,labelfont=sf,textfont=sf]{subfig}
\usepackage{textcomp}
\usepackage{stfloats}
\usepackage{url}
\usepackage{verbatim}
\usepackage{graphicx}
\usepackage{cite}
\usepackage{adjustbox}
\usepackage{multirow}
\usepackage{booktabs}
\usepackage{algorithm}
\usepackage{hyperref}
\usepackage{pifont}
\usepackage{dsfont}
\usepackage{tikz,xcolor}

\hypersetup{hidelinks,
	colorlinks=true,
	allcolors=black,
	pdfstartview=Fit,
	breaklinks=true}
	
\definecolor{lime}{HTML}{A6CE39}
\DeclareRobustCommand{\orcidicon}{
\begin{tikzpicture}
\draw[lime, fill=lime] (0,0)
circle[radius=0.16]
node[white]{{\fontfamily{qag}\selectfont \tiny \.{I}D}};
\end{tikzpicture}
\hspace{-2mm}
}
\foreach \x in {A, ..., Z}{%
\expandafter\xdef\csname orcid\x\endcsname{\noexpand\href{https://orcid.org/\csname orcidauthor\x\endcsname}{\noexpand\orcidicon}}
}

\begin{document}

\title{Multi-QuAD: Multi-Level Quality-Adaptive Dynamic Network for Reliable Multimodal Classification}


\author{Shu Shen\hspace{-1.5mm}\orcidB{},~\IEEEmembership{Student Member,~IEEE,} C. L. Philip Chen\hspace{-1.5mm}\orcidD{},~\IEEEmembership{Life Fellow,~IEEE, } and Tong Zhang}

\markboth{Journal of \LaTeX\ Class Files,~Vol.~14, No.~8, May~2025}%
{Shell \MakeLowercase{\textit{et al.}}: A Sample Article Using IEEEtran.cls for IEEE Journals}


\maketitle

\begin{abstract}
Multimodal machine learning has achieved remarkable progress in many scenarios, but its reliability is undermined by varying sample quality. This paper finds that existing reliable multimodal classification methods not only fail to provide robust estimation of data quality, but also lack dynamic networks for sample-specific depth and parameters to achieve reliable inference. To this end, a novel framework for multimodal reliable classification termed \textit{Multi-level Quality-Adaptive Dynamic multimodal network} (Multi-QuAD) is proposed. Multi-QuAD first adopts a novel approach based on noise-free prototypes and a classifier-free design to reliably estimate the quality of each sample at both modality and feature levels. It then achieves sample-specific network depth via the \textbf{\textit{Global Confidence Normalized Depth (GCND)}} mechanism. By normalizing depth across modalities and samples, \textit{\textbf{GCND}} effectively mitigates the impact of challenging modality inputs on dynamic depth reliability. Furthermore, Multi-QuAD provides sample-adaptive network parameters via the \textbf{\textit{Layer-wise Greedy Parameter (LGP)}} mechanism driven by feature-level quality. The cross-modality layer-wise greedy strategy in \textbf{\textit{LGP}} designs a reliable parameter prediction paradigm for multimodal networks with variable architecture for the first time. Experiments conducted on four datasets demonstrate that Multi-QuAD significantly outperforms state-of-the-art methods in classification performance and reliability, exhibiting strong adaptability to data with diverse quality.
\end{abstract}


\section{Introduction}\label{sec:introduction}

\begin{figure}[!t] 
    \centering 
    \subfloat[]{ 
        \includegraphics[width=\linewidth]{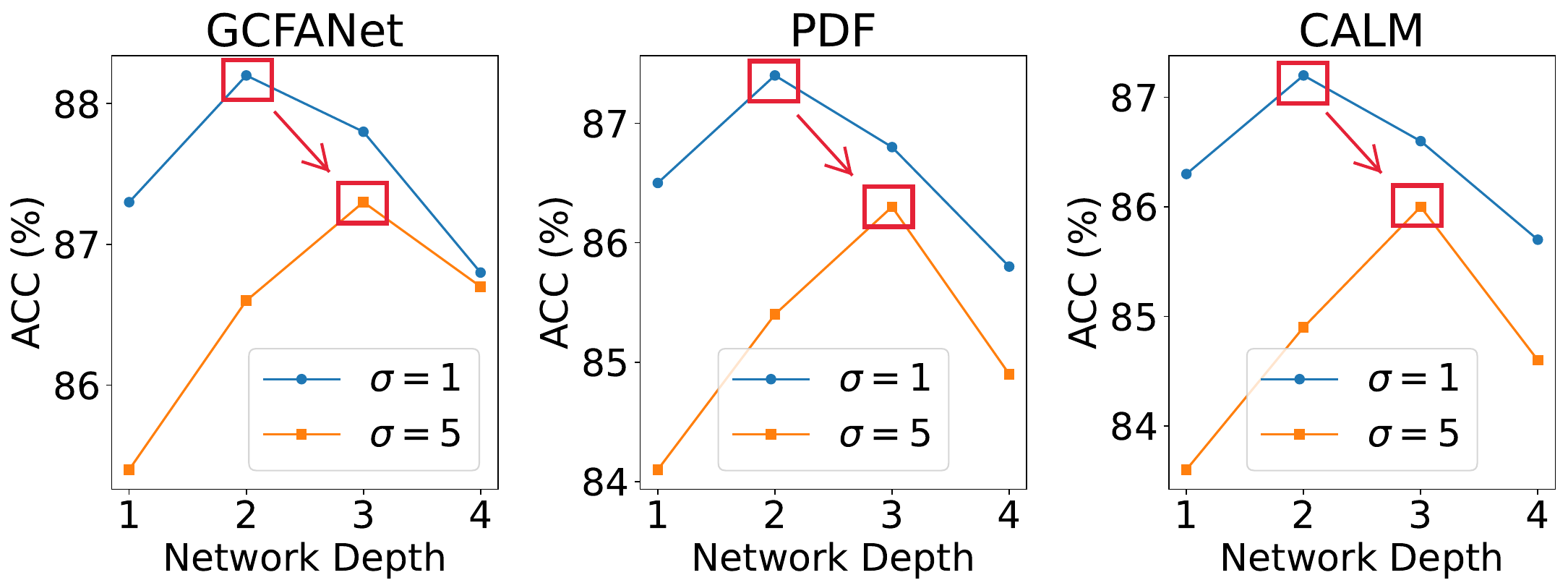}
        \label{fig:1a}
    }
    \vspace{0.000001cm} 
    \subfloat[]{ 
        \includegraphics[width=0.47\linewidth]{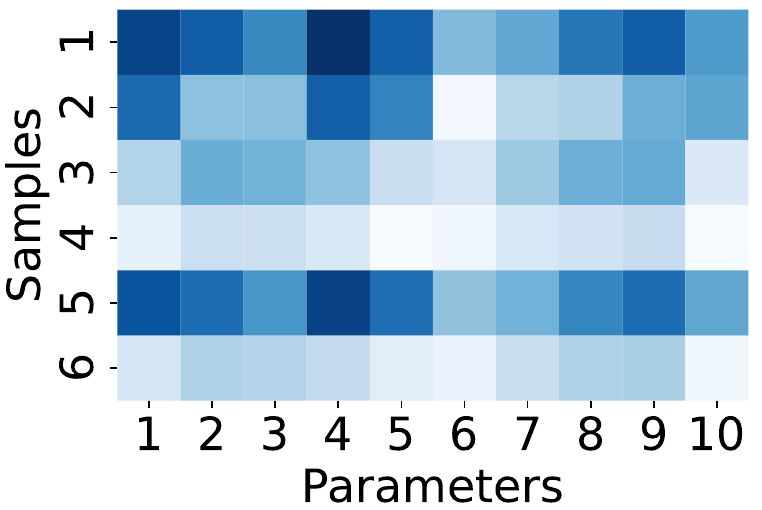}
        \label{fig:1b}
    }
    \hfill 
    \subfloat[]{ 
        \includegraphics[width=0.47\linewidth]{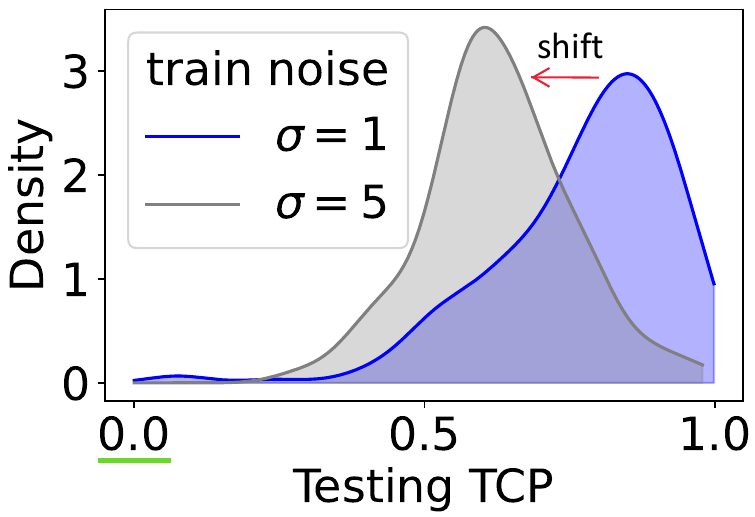}
        \label{fig:1c}
    }
    \caption{Empirical studies under varying data quality. We simulate data quality degradation by adding Gaussian noise to one of the modalities on BRCA dataset, with $\sigma$ representing the noise intensity. (a) Classification accuracy (ACC) of three state-of-the-art methods (GCFANet \protect\cite{zheng2024global}, PDF \protect\cite{pmlr-v235-cao24c}, CALM \protect\cite{zhou2023calm}) under different depths of their unimodal network depths corresponding to the noisy modality. (b) Visualization of the model parameters required to map samples with different features to their corresponding class centers. (c) The confidence estimation results on the \textit{\textbf{same test samples}} provided by TCP \protect\cite{han2022multimodal} as a representative example after \textit{\textbf{training on data with different noise intensities.}} \textbf{More details of the observation experiments are included in the Supplemental Materials.}}
    \label{fig:group}
\end{figure}

\IEEEPARstart{T}{he} accessibility of multimodal data has significantly enhanced the performance of machine learning in various scenarios \cite{dou2022empirical,li2022clip,sun2022cubemlp,elizalde2023clap,10339893,10171388,9681296,10814984,10577436,9863920,10076804,10445009}, especially in multimedia applications \cite{10950092,10980423,10980361}. 
However, multimodal samples often exhibit complex quality variations, which are primarily attributed to the inherent differences in the informativeness across modalities \cite{peng2022balanced,ma2023calibrating,wei2024fly} and the noise introduced by varying sensor conditions or environmental factors \cite{cheng2019noise,geng2021uncertainty,zhang2024multimodal}. This phenomenon poses substantial challenges for reliable model prediction, especially in safety-critical domains such as computer-aided diagnosis and autonomous driving \cite{feng2018towards,nair2020exploring,han2022multimodal,zhou2023calm}. To tackle this challenge, extensive research has explored the variability in multimodal sample quality at both modality and feature levels \cite{geng2021uncertainty,han2022multimodal,zou2023dpnet,zheng2023multi,tellamekala2023cold,zhou2023calm,zhang2023provable,zheng2024global,pmlr-v235-cao24c}, significantly enhancing classification reliability by removing low-quality and uninformative features. The modality level refers to the quality variations of modalities among samples, while the feature level pertains to the quality differences of individual feature values within each modality's feature vector across samples.

However, empirical studies reveal several limitations of current reliable multimodal classification methods when dealing with data of varying quality. \textbf{(1) Lack of network dynamics.} We introduced Gaussian noise to one modality of the data to simulate modality-level quality degradation and implemented versions of the state-of-the-art model with different combinations of unimodal network depths. Fig. \ref{fig:1a} shows that as the quality degrades (increasing noise intensity $\sigma$ from 1 to 5), the unimodal networks corresponding to the affected modality in all SOTA models consistently require greater depth to achieve the optimal accuracy. Additionally, the experimental results shown in Fig. \ref{fig:1b} indicate that different network parameters are required to map feature vectors of samples with inherent feature-level quality differences to the most reliable representation (e.g., their corresponding class centers). Unfortunately, existing methods are static networks with fixed network depth and parameters. \textbf{(2) Lack of reliable confidence estimation.} These methods typically provide confidence estimation results that overfit the training noise, leading to decreased reliability. As shown in Fig. \ref{fig:1c} and \ref{fig_compare_conf}, existing confidence estimation methods yield significantly different results on the same test samples after being trained on data with different noise settings. Moreover, they cannot simultaneously perform quality estimation of a sample at both modality and feature levels. Thus, we pose two key questions: \textbf{\textit{(1) How can we design a robust multi-level quality estimation method to support reliable network adjustments? (2) How can we adjust the network's depth and parameters to ensure reliability for a given multimodal sample?}}

This paper addresses the above questions by proposing Multi-level Quality-Adaptive Dynamic multimodal network (Multi-QuAD), a novel framework for reliable multimodal classification. Multi-QuAD first adopts \textbf{\textit{noise-free prototype confidence estimation (NFCE)}} to reliably estimate sample quality at both the modality and feature levels. Then, Multi-QuAD sequentially employs the \textit{\textbf{Global Confidence Normalized Depth (GCND)}} and \textbf{\textit{Layer-wise Greedy Parameter (LGP)}} mechanisms to adjust its depth and parameters based on modality- and feature-level qualities of the sample, respectively. Unlike existing confidence estimation methods, \textbf{\textit{NFCE}} derives class probability via a classifier-free design using noise-free prototypes (noise-removed class centers) pre-optimized based on inter-modality and inter-class relationships, instead of using network structures or classifiers. This design not only enables multi-level quality estimation but also facilitates reliable quality estimation by avoiding overfitting to training noise in the network structure or classifier. The \textbf{\textit{GCND}} mechanism maps sample modality quality to network depth. Notably, it reduces the impact of extremely difficult modality inputs by normalizing depths across modalities of all samples, thereby enhancing the reliability of dynamic depth. The \textbf{\textit{LGP}} mechanism further performs cross-modality parameters prediction driven by a novel shallow-to-depth layer-wise greedy strategy aiming to maximize feature-level quality enhancement at each network layer. This mechanism introduces an efficient and reliable parameter prediction framework on a flexible multimodal framework with dynamic depths for the first time.

The contributions of this paper can be summarized as:
\begin{itemize}
\item We propose Multi-QuAD, which introduces a novel perspective on reliable multimodal classification by dynamically adapting network depth and parameters to varying data quality at both modality and feature levels. Multi-QuAD significantly outperforms state-of-the-art methods in accuracy and reliability across multiple benchmarks.
\item The proposed classifier-free method \textbf{\textit{NFCE}} leverages noise-free prototypes pre-optimized by inter-modality and inter-class relations, outperforming existing confidence estimation methods in avoiding noise overfitting and enabling multi-level quality assessment.
\item The two proposed dynamic mechanisms, \textbf{\textit{GCND}} and \textbf{\textit{LGP}}, with cross-modality depth normalization and layer-wise greedy parameter prediction, design a new paradigm that endows dynamic networks with multimodal reliability for the first time.
\end{itemize}


\section{Related Works}
This section reviews some related works on reliable multimodal classification, classification confidence estimation, and dynamic neural networks.

\subsection{Reliable Multimodal Classification}
Multimodal learning has achieved significant success in enhancing model performance across numerous application fields \cite{10058603,9930669,9906434,9121752,9246699,10313078,10214099,10374242}. However, multimodal data often contains noise, leading to the emergence of low-quality data, which in turn affects the reliability of multimodal learning. To this end, many studies have developed quality-aware methods to provide reliable multimodal classification results from low-quality multimodal data. Some methods have focused on addressing modality-level quality issues. For example, Han et al. \cite{han2020trusted} have parameterized the evidence of different modality features using Dirichlet distribution and fused each modality based on their evidence using Dempster-Shafer theory. Geng et al. \cite{geng2021uncertainty} have proposed the DUA-Nets, which achieved multimodal representation learning based on modality uncertainty. Zhang et al. \cite{zhang2023provable}, Cao et al. \cite{pmlr-v235-cao24c} have achieved more robust and generalized multimodal fusion by dynamically assigning weights to each modality based on their uncertainty. In contrast, some studies address data quality issues across multiple levels. For instance, Han et al. \cite{han2022multimodal} have modeled informativeness at the feature and modality levels, achieving trustworthy multimodal feature fusion. Zhou et al. \cite{zhou2023calm} have learned trustworthy features in each modality and achieved confidence-aware fusion among modalities. Despite their effectiveness, these methods are all based on static networks, whose limitations have been analyzed in Sec. \ref{sec:introduction}. To address this, we propose Multi-QuAD, which possesses network dynamics and adaptability towards different input samples.

\subsection{Confidence Estimation}
In reliable multimodal classification methods, confidence estimation can reflect the quality of the model input, serving as a necessary foundation for the model's reliability. In the early years, Bayesian neural networks \cite{denker1990transforming,mackay1992bayesian,neal2012bayesian}, predictive confidence \cite{hendrycks2016baseline}, and Dempster-Shafer theory \cite{han2020trusted} were proposed for confidence estimation. Predictive confidence measures the consistency between the model's predicted class probabilities and its empirical accuracy, which typically refers to the accuracy in classification tasks. Dempster-Shafer theory (DST) measures uncertainty by assigning probabilities to sets rather than individual events, allowing for handling information at different precision levels without introducing additional assumptions. In recent years, several easily implementable, lightweight confidence estimation methods have been proposed. For example, Maximum-Class Probability (MCP) \cite{zou2023dpnet} uses the maximum value of the classification probability predicted by the model as the confidence assessment. However, MCP suffers from significant overconfidence issues \cite{liu2020energy,han2022multimodal}. To address this, methods such as energy score \cite{zhang2023provable} and True-Class-Probability (TCP) \cite{han2022multimodal,pmlr-v235-cao24c} have been proposed, which leverage the model's predicted probability to obtain more reliable confidence assessments. However, these methods, which utilize classification probabilities predicted by network architectures or classifiers, suffer from overfitting to training noise and are incapable of multi-level quality assessment. To address these common issues, we propose noise-free prototype confidence estimation (\textbf{\textit{NFCE}}) based on noise-free prototypes and a classifier-free design.

\begin{figure*}[t]
\centering
\includegraphics[width=0.9\textwidth]{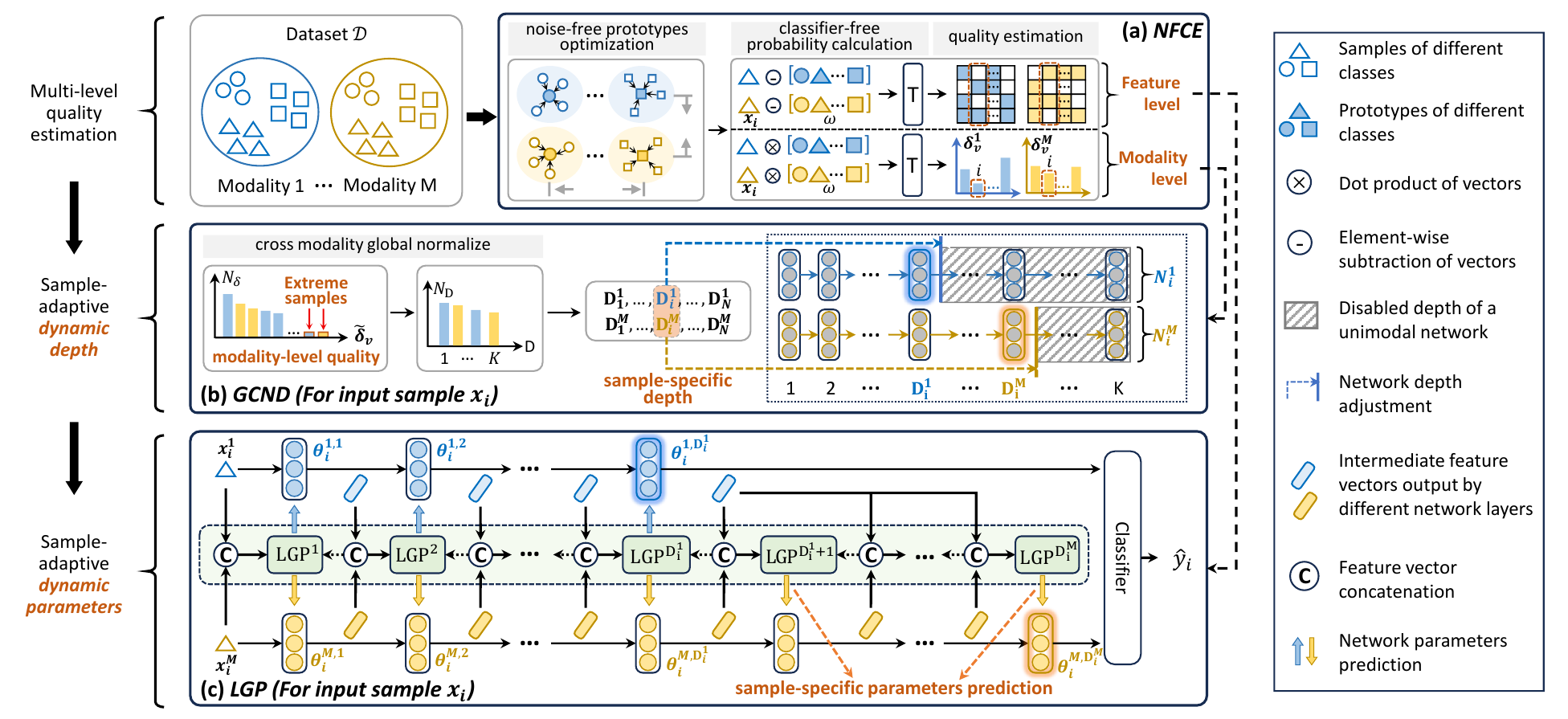} 
\caption{The framework of the proposed Multi-QuAD (better viewed in colour). Without loss of generality, this figure illustrates the case of two modalities, with blue and yellow representing different modalities. For an input sample $x_i$ from dataset $\mathcal{D}$: (a) the modality-level and feature-level quality of $x_i$ is estimated via \textbf{\textit{Noise-free Prototype Confidence Estimation (NFCE)}}. (b) The reliable depth of Multi-QuAD for $x_i$ is adjusted by \textbf{\textit{Global Confidence Normalized Depth (GCND)}} based on modality-level quality. (c) The reliable parameters of Multi-QuAD for $x_i$ are adjusted by \textbf{\textit{Layer-wise Greedy Parameter (LGP)}} driven by feature-level quality. The detailed implementation of $\textbf{LGP}^t$ at each layer will be demonstrated in Fig. \ref{fig:lgp}.}
\label{fig_model_arch}
\end{figure*}

\subsection{Dynamic Neural Network}
Designing dynamic architectures or parameters is a promising approach to promoting neural networks' accuracy, computational efficiency, and adaptability in multiple tasks \cite{veit2018convolutional,hua2019channel,su2019pixel,han2021dynamic}. In image classification tasks, several works \cite{huang2017multi,wang2018skipnet,jie2019anytime} have utilized the concept of early exit to enable efficient inference for samples. 
In natural language processing tasks, many studies \cite{xin2020deebert,zhou2020bert} have effectively improved inference speed and addressed the overthinking issue by allowing samples to be output at different network layers. In object detection tasks, Lin et al. \cite {lin2023dynamicdet} have achieved accuracy-speed trade-offs by applying different network routines to different samples. In the multimodal task, DynMM \cite{xue2023dynamic} has employed a dynamic fusion strategy to achieve high computational efficiency in emotion recognition. However, dynamic neural networks (DNNs) generally lack multimodal reliability. Most DNNs, except for DynMM \cite{xue2023dynamic}, are unimodal and cannot be extended to multimodal scenarios. All DNNs, including DynMM, commonly lack reliability in both confidence estimation and dynamic mechanism design. The proposed Multi-QuAD complements these methods by offering reliable multi-level confidence estimation and robust dynamic mechanisms based on the sample's quality and cross-modality relationships, which endows DNNs with multimodal reliability for the first time.

\section{Proposed Approach}
In this section, we provide a detailed illustration of the proposed Multi-QuAD. Given a multimodal dataset $\mathcal{D}=\{(x_i,y_i)\}_{i=1}^N$ containing $N$ samples with $M$ modalities belonging to $C$ classes. Multimodal classification methods aim to learn a neural network that maps every input sample $x_i=\{x_i^m\in\mathbb{R}^{d^m}\}_{i=m}^M$ to its corresponding class label $y_i\in\mathbb{R}^C$, where $d^m$ is the dimension of the $m$-th modality. In this work, we propose Multi-QuAD to provide reliable classification result $\hat{y}_i$ for input sample $x_i$ via sample-adaptive network depth and parameters.



\subsection{Backbone of Multi-QuAD} \label{sec:backbone}

We first introduce the hierarchical backbone structure of Multi-QuAD. Multi-QuAD is an adaptive network that dynamically adjusts according to different input samples. It consists of $M$ unimodal networks, each formed by stacking an adjustable number of network blocks. Each block is composed of a single neural network layer, with the parameters dynamically predicted based on the input sample. Given an input sample $x_i=\{x_i^m\}_{m=1}^M$, the adjusted Multi-QuAD is denoted as $QUAD_i$, where each unimodal network is denoted as $N_i^m (m\in [1,M])$. The depth of each unimodal network $N_i^m$ is adjusted to $\textbf{D}_i^m$ based on the input $x_i$, i.e., stacking $\textbf{D}_i^m$ network blocks in $N_i^m$, as elaborated in Sec. \ref{sec:dydepth}. The $t$-th $(t\in [1,\textbf{D}_i^m])$ network block in $N_i^m$ is denoted as $B_i^{m,t}$, and its network layer parameters are denoted as $\theta_i^{m,t}$. $\theta_i^{m,t} = [W_i^{m,t}, b_i^{m,t}]$ where $W_i^{m,t}$ and $b_i^{m,t}$ represent the weight and bias, respectively. The parameters $\theta_i^{m,t}$ of each block are dynamically predicted during the inference process for the sample $x_i$, as elaborated in Sec. \ref{sec:dyparam}.
Input feature vectors of each modality in $x_i$ are gradually enhanced through network blocks in their respective unimodal networks, and finally obtain high-quality output features $f_i=\{f_i^m\}_{m=1}^M$:
\begin{align}
    f_i^m&=N^m_i(x_i^{m})=B^{m,\textbf{D}_i^m}_i\circ...\circ B^{m,1}_i(x_i^m),
\end{align}
where $\circ$ refers to the composition of two operations. The calculation in the network block $B^{m,t}_i$ can be formulate as:
\begin{align}
    q^{m,t}_i=\sigma(W_i^{m,t}\cdot I^{m,t}_i+b_i^{m,t}) \notag \\
    I^{m,t+1}_i=q^{m,t}_i\odot I^{m,t}_i.
    \label{eq:q-enhance}
\end{align}
$I^{m,t}_i$ and $I^{m,t+1}_i$ denote the input and output of the network block $B^{m,t}_i$, respectively, and $I^{m,1}_i=x_i^m, I^{m,\textbf{D}_i^m+1}_i=f_i^m$.
$\sigma$ is the sigmoid function. $q_i^{m,t}\in\mathbb{R}^{d^m}$ is the feature-level informativeness vector of $I_i^{m,t}$ learned by $B^{m,t}_i$, with each element reflects the informativeness of each feature value in $I^{m,t}_i$. The quality enhancement is achieved by retaining informative features while removing redundant ones through element-wise multiplication $\odot$ between $q^{m,t}_i$ and $I^{m,t}_i$.

The output high-quality feature vectors of all modalities $\{f_i^{m}\}_{m=1}^M$ are finally concatenated to provide the reliable classification result $\hat{y}_i$ of the input sample $x_i$. The overall framework is trained by minimizing the classification loss $\mathcal{L}^{task}$, i.e., the cross-entropy loss between predictions $\{\hat{y}_i\}_{i=1}^N$ and ground-truth labels $\{y_i\}_{i=1}^N$.

\begin{figure}[t]
\centering
\includegraphics[width=0.95\linewidth]{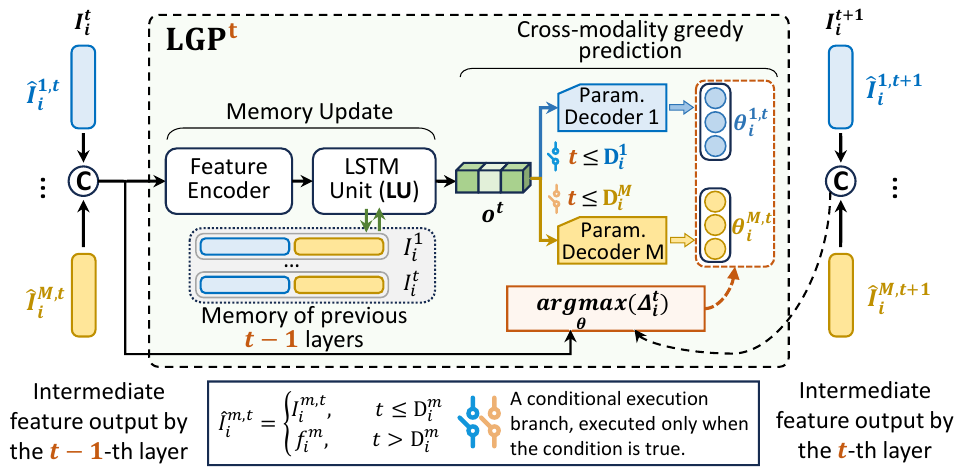} 
\caption{The detailed implementation of cross-modality greedy parameter prediction at the $t$-th layer ($\textbf{LGP}^t$).}
\label{fig:lgp}
\end{figure}

\subsection{Noise-free Prototype Confidence Estimation} \label{sec:nfce}
This section introduces the Noise-free Prototype Confidence Estimation (\textbf{\textit{NFCE}}) based on pre-optimized noise-free prototypes and classifier-free design.

\subsubsection{Noise-Free Prototype Pre-optimization}
For the dataset $\mathcal{D}$ ($M$ modalities and $C$ classes), we aim to find $C\times M$ feature vectors $\omega=\{\{\omega^{c,m}\}_{c=1}^C\}_{m=1}^M$ termed noise-free prototypes. $\omega^{c,m}$ is a noise-free high-confidence representative of the feature vectors of the $m$-th modality from all samples under the $c$-th class. To learn $\omega$, feature vectors of all input samples are first enhanced by modality-specific encoders. Then, the mean of all enhanced feature vectors of different modalities under different classes is calculated and yields $C\times M$ prototypes $\mu=\{\{\mu^{c,m}\}_{c=1}^C\}_{m=1}^M$. Subsequently, we eliminate noise and enhance the classification confidence of these prototypes $\mu$, optimizing them into $\omega$ by maximizing the constraint $\mathcal{L}^{rob}$: $\omega=\mathop{\arg\max}\limits_{\mu} \mathcal{L}^{rob}$, where $\mathcal{L}^{rob}$ is defined as:
\begin{align}
    \mathcal{L}^{rob} =\frac{1}{CM^2}\sum_{c=1}^C{\sum_{i=1}^M{\sum_{j=1,j\neq i}^M{I(\mu^{c,i},\mu^{c,j})}}} \notag \\
    +\frac{1}{CM}\sum_{m=1}^M{\sum_{c=1}^C{log(\frac{exp(cos(\mu^{c,m},\mu^{c,m}))}{\sum_{i=1}^C{exp(cos(\mu^{c,m},\mu^{i,m}))}})}}.
    \label{eq:rob}
\end{align}
$cos(\cdot,\cdot)$ is the cosine similarity. $I(\cdot,\cdot)$ refers to mutual information between the two representations. Maximizing the first term of $\mathcal{L}^{rob}$ removes noise and redundant features by increasing the shared information between modalities \cite{federici2020learning}. Maximizing the second term improves classification confidence by encouraging orthogonality among classes \cite{huang2020deep}.

\subsubsection{Multi-level Reliable Quality Estimation}\label{sec:ml}
Instead of using network structures or classifiers, a classifier-free design is applied that obtains noise-reliable class probability at both modality and feature levels for each input sample $x_i=\{x_i^m\}_{m=1}^M$ by contrasting it with pre-optimized noise-free prototypes $\omega$ at different levels. The modality-level class probability $p_v$ is obtained by computing the cosine similarities between the feature vector $x_i^m$ of each modality $m$ and all noise-free prototypes $\omega^m=[\omega^{c,m}]_{c=1}^C$ of corresponding modality. For the feature-level probability $p_f$, the feature-wise absolute difference between $x_i^m$ and $\omega^m$ is calculated:
\begin{align}
p_{v}(x_i^m)&=Softmax([cos(x_i^{m},\omega^{c,m})]_{c=1}^C)\in\mathbb{R}^C, \label{eq:conf-m} \\
p_{f}(x_i^m)&=Softmax([|x_i^m-\omega^{c,m}|]_{c=1}^C)\in\mathbb{R}^{d^m\times C}.
\end{align}
After the reliable class probabilities $p_v$ and $p_f$ are obtained, the modality-level quality $\delta_v$ and feature-level quality $\delta_f$ are estimated using the true class probability $\mathcal{T}$ \cite{han2022multimodal}: 
\begin{align}
    \delta_v(x_i^m)=\mathcal{T}(p_v(x_i^m)) \in [0,1], \label{eq:modality-level-quality}\\
    \delta_f(x_i^m)=\mathcal{T}(p_{f}(x_i^m)) \in\mathbb{R}^{d^m}. \label{eq:feature-level-quality}
\end{align}
\textbf{The detailed implementation of $\mathcal{T}$ is given in the Supplemental Materials.}

\begin{algorithm}[t]
\caption{The layer-wise greedy parameter (\textbf{\textit{LGP}}) mechanism of Multi-QuAD}
\label{alg:algorithm}
\textbf{Input}: Multimodal input samples $\{x_i\}_{i=1}^N=\{x_i^1,...,x_i^M\}_{i=1}^N$, the depths of Multi-QuAD corresponding to each sample $\{\textbf{D}_i\}_{i=1}^N=\{\textbf{D}_i^1,...\textbf{D}_i^M\}_{i=1}^N$ provided by the \textbf{\textit{GCND}} mechanism. The LSTM unit $\textbf{LU}$. The hidden state dimension number $hdim$. Modality-specific parameter converters $\{\textbf{PD}^m\}_{m=1}^M$. The feature-level quality estimation function $\delta_f(\cdot)$. Random tensor generating function $RT(\cdot)$.

\textbf{Output}: The parameters of each network block of Multi-QuAD for each sample: $\theta_i^{m,t}$, reliable classification result of each sample: $\hat{y}_i$. The total feature-level quality gain $\Delta$.

\begin{algorithmic}[1] 
\STATE $\Delta\leftarrow 0$
\FOR{each sample $x_i=\{x_i^1,...,x_i^M\}$}
\STATE Depths of all unimodal networks in Multi-QuAD $\textbf{D}_i=\{\textbf{D}_i^1,...,\textbf{D}_i^M\}$.
\STATE The input feature vectors of all modalities at a certain depth: $\{\hat{I}_i^{1},...,\hat{I}_i^{M}\}$.
\STATE The state of the $\textbf{LU}$ at a certain depth: $(h,c)$.
\FOR{$t=1$ to $max(\textbf{D}_i^1,...,\textbf{D}_i^M)$}
\IF{$t=1$}
\STATE $\{\hat{I}_i^{1},...,\hat{I}_i^{M}\}\leftarrow \{x_i^1,...,x_i^M\}$.
\STATE $(h,c)\leftarrow(RT(hdim),RT(hdim))$.
\ENDIF
\STATE $I_i\leftarrow concat(\hat{I}_i^{1},...,\hat{I}_i^{M})$.
\STATE $\textbf{o}^t,(h,c)\leftarrow\textbf{LU}(I_i,(h,c))$.
\FOR{$m=1$ to $M$}
\IF{$t\leq \textbf{D}_i^m$}
\STATE $\theta_i^{m,t}=[W_i^{m,t},b_i^{m,t}]\leftarrow\textbf{PD}^m(\textbf{o}^t)$
\STATE $\hat{I}_i^{m+1}\leftarrow\sigma(W_i^{m,t}\cdot \hat{I}_i^{m}+b_i^{m,t})\odot \hat{I}_i^{m}$
\STATE $\Delta\leftarrow\Delta+mean(max(\delta_f(\hat{I}_i^{m+1})-\delta_f(\hat{I}_i^m),\textbf{0}))$
\ENDIF
\ENDFOR
\ENDFOR
\STATE The high-quality output feature vectors of sample $x_i$: $f_i=\{f_i^1,...f_i^M\}\leftarrow \{\hat{I}_i^1,...,\hat{I}_i^M\}$.
\STATE The reliable classification result of $x_i$: $\hat{y}_i\leftarrow Classifier(concat(f_i^1,...f_i^M))$.
\ENDFOR
\end{algorithmic}
\label{alg:lgp}
\end{algorithm}

\subsection{Global Confidence Normalized Depth}\label{sec:dydepth}
For input sample $x_i$, the network depth of Multi-QuAD is adjusted to $\{\textbf{D}_i^m\}_{m=1}^M$ based on its modality-level qualities $\{\delta_v(x_i^m)\}_{m=1}^M$. Based on our observation in Fig. \ref{fig:1a}, the lower-quality modality requires a deeper unimodal network to enhance its features. Thus, a positive correlation must be maintained between $\Tilde{\delta}_{v,i}^m=1-\delta_v(x_i^m)$ and the network depths $\textbf{D}_i^m$, where $\delta_v(x_i^m)$ is the quality of the modality $m$ input of $x_i$ estimated by \textbf{\textit{NFCE}}. 
Instead of adjusting the depth of each modality for each sample individually, Multi-QuAD employs a global cross-modality depth normalization strategy. This strategy simultaneously normalizes $\{\{\Tilde{\delta}_{v,i}^m\}_{m=1}^M\}_{i=1}^N$ including all modalities from all samples into a unified integer space. The normalized values serve as the corresponding network depths of each modality for each sample $\textbf{D}=\{\{\textbf{D}_i^m\}_{m=1}^M\}_{i=1}^N$. On the one hand, this strategy achieves a positive correlation mapping. On the other hand, it offers a simple yet effective way to enhance the reliability of dynamic depth by normalizing the network depth distribution across modalities and mitigating the adverse effects of extreme modality samples. The calculation of this strategy can be formulated as:
\begin{align}
    \textbf{D}=\{\{\textbf{D}_i^m\}_{m=1}^M\}_{i=1}^N=\mathcal{N}_K(\{\{\Tilde{\delta}_{v,i}^m\}_{m=1}^M\}_{i=1}^N).
\end{align}
$\mathcal{N}_K(\cdot)$ is a function that scales all the input values proportionally to integers within the range $[1,K]$, where $K\geq 1$ is an integer hyperparameter. The analysis of $K$ is provided in Sec. \ref{sec:hyper-param-analy}.

\begin{table*}[h]
\caption{Comparison with state-of-the-art reliable multimodal classification methods on four datasets.}
\begin{center}
\begin{tabular}{@{}c|c|ccc|ccc@{}}
\toprule
       \multirow{2}{*}{Method} & \multirow{2}{*}{Type}   & \multicolumn{3}{c|}{BRCA}   & \multicolumn{3}{c}{ROSMAP}    \\ 
 & & ACC & WeightedF1 & MacroF1 & ACC & F1 & AUC \\ 
 \midrule 
MD \cite{han2022multimodal} &static&  87.7$\pm$0.3 & 88.0$\pm$0.5 & 84.5$\pm$0.5       & 84.2$\pm$1.3  & 84.6$\pm$0.7  & 91.2$\pm$0.7       \\ 
MLCLNet \cite{zheng2023multi}&static& 86.4$\pm$1.6& 87.8$\pm$1.7& 82.6$\pm$1.8      & 84.4$\pm$1.5   & 85.2$\pm$1.5       & 89.3$\pm$1.1  \\ 
DPNET \cite{zou2023dpnet}&static& 87.8$\pm$1.0& 88.4$\pm$1.2& 85.2$\pm$1.2       & 85.1$\pm$1.1   & 84.8$\pm$0.7         & 91.3$\pm$0.7      \\ 
CALM \cite{zhou2023calm}&static& 88.2$\pm$0.7& 88.5$\pm$0.8& 85.1$\pm$0.8       & 85.5$\pm$1.2   & 87.9$\pm$0.9         & 91.3$\pm$1.0      \\ 

QMF \cite{zhang2023provable}&static& 87.4$\pm$0.4& 87.7$\pm$0.5& 84.1$\pm$0.5       & 84.6$\pm$0.9   & 84.8$\pm$0.8         & 90.5$\pm$0.8      \\ 
GCFANet \cite{zheng2024global}&static& \underline{88.6$\pm$1.5} & \underline{88.9$\pm$1.6} & \underline{85.3$\pm$1.6}     & \underline{86.3$\pm$1.4} & \underline{88.3$\pm$1.6}        & 91.5$\pm$1.2      \\ 
PDF \cite{pmlr-v235-cao24c}&static& 88.2$\pm$0.7 & 88.0$\pm$0.6 & 84.9$\pm$0.6     & 85.9$\pm$0.5 & 88.0$\pm$0.4        & \underline{91.6$\pm$0.5}  \\
DynMM \cite{xue2023dynamic}&dynamic& 82.3$\pm$1.7 & 82.4$\pm$1.7 & 80.1$\pm$1.2     & 81.1$\pm$1.5 & 80.2$\pm$1.6        & 80.4$\pm$1.3      \\
\midrule
Multi-QuAD  & dynamic&\textbf{93.4$\pm$ 0.3} & \textbf{93.5$\pm$ 0.4} & \textbf{91.5$\pm$ 0.4} & \textbf{92.2$\pm$ 0.4} & \textbf{92.7$\pm$ 0.3} & \textbf{95.2$\pm$ 0.3} \\ \bottomrule
\toprule
\multirow{2}{*}{Method} & \multirow{2}{*}{Type} \  & \multicolumn{3}{c|}{CUB}   & \multicolumn{3}{c}{FOOD101}    \\ 
& & ACC & WeightedF1 & MacroF1 & ACC & WeightedF1 & MacroF1 \\ \midrule 
MD \cite{han2022multimodal} &static&  90.1$\pm$0.7 & 90.0$\pm$0.8 & 89.9$\pm$0.7       & 92.8$\pm$0.3  & 92.6$\pm$0.2  & 92.6$\pm$0.2       \\ 
MLCLNet \cite{zheng2023multi}&static& 88.2$\pm$1.4& 88.3$\pm$1.7& 87.9$\pm$1.3      & 92.1$\pm$1.0   & 92.2$\pm$1.2       & 92.1$\pm$1.1  \\ 
DPNET \cite{zou2023dpnet}&static& 92.1$\pm$0.9& 91.8$\pm$0.7& 92.2$\pm$1.1       & 93.1$\pm$1.2   & 93.0$\pm$0.7         & 92.8$\pm$0.8      \\ 
CALM \cite{zhou2023calm}&static& 92.0$\pm$0.2& 92.1$\pm$0.4& 92.1$\pm$0.5       & 93.0$\pm$1.1   & 92.9$\pm$0.7         & 92.9$\pm$0.8      \\ 
QMF \cite{zhang2023provable}&static & 89.8$\pm$0.4& 89.7$\pm$0.5& 89.1$\pm$0.5       & 92.7$\pm$0.5   & 92.3$\pm$0.4         & 92.4$\pm$0.4      \\ 
GCFANet \cite{zheng2024global}&static& 92.3$\pm$1.2 & 92.4$\pm$1.1 & 92.6$\pm$1.2     & 92.9$\pm$1.2 & 93.1$\pm$1.1        & 92.7$\pm$0.9      \\ 
PDF \cite{pmlr-v235-cao24c}&static& \underline{93.0$\pm$0.5} & \underline{93.2$\pm$0.4} & \underline{93.2$\pm$0.4}     & \underline{93.3$\pm$0.6} & \underline{93.5$\pm$0.7}       & \underline{93.0$\pm$0.6}  \\ 
DynMM \cite{xue2023dynamic}&dynamic& 88.3$\pm$1.5 & 88.3$\pm$1.6 & 87.7$\pm$1.5    & 89.3$\pm$1.6 & 89.4$\pm$1.7       & 88.5$\pm$1.5      \\
\midrule
Multi-QuAD  & dynamic&\textbf{97.4$\pm$ 0.2} & \textbf{97.3$\pm$ 0.3} & \textbf{97.4$\pm$ 0.4} & \textbf{94.5$\pm$ 0.3} & \textbf{94.7$\pm$ 0.4} & \textbf{94.4$\pm$ 0.3} \\ \bottomrule
\end{tabular}
\end{center}
\label{sota table1}
\end{table*}

\subsection{Layer-wise Greedy Parameter} \label{sec:dyparam}
After the depths $\{\textbf{D}_i^m\}_{m=1}^M$ of Multi-QuAD for the sample $x_i$ is determined, the Layer-wise Greedy Parameter (\textbf{\textit{LGP}}) mechanism predicts the optimal network parameters for all network blocks in Multi-QuAD. To achieve efficient and reliable parameter prediction on a flexible multimodal framework with dynamic depths, a shallow-to-deep layer-wise greedy prediction strategy is proposed. Specifically, at each depth, the greedy strategy predicts the optimal network parameters jointly across all unimodal networks, aiming to maximize the feature-level quality enhancement in the current layer based on how the features have been enhanced in previous depths. For a certain depth $t\in[1,max(\textbf{D}_i^1,...,\textbf{D}_i^M)]$, denote the ``latest'' output feature vectors from each unimodal network as $\hat{I}_i^{m,t}$:
\begin{align}
\hat{I}_i^{m,t} =
\begin{cases}
I_i^{m,t}, & \text{if } t \leq \textbf{D}_i^m \\
f_i^m, & \text{otherwise}.
\end{cases}
\end{align}
$I_i^{m,t}$ is the input feature vector of $B_i^{m,t}$ output by $B_i^{m,t-1}$, and $f_i^m$ is the feature vector output by the final layer of $N_i^m$, as mentioned in Sec. \ref{sec:backbone}. Let $I_i^t=Concat(\{\hat{I}_i^{m,t}\}_{m=1}^M)$. The greedy strategy $\textbf{LGP}^t$ at the $t$-th layer takes the enhanced feature vectors $I_i^1, \ldots, I_i^t$ output by the previous $t-1$ layers into account and jointly predicts the parameter values for the $t$-th block of all unimodal networks, such that the feature-level quality gain $\Delta_i^t$ at depth $t$ is maximized. The mechanism of $\textbf{LGP}^t$ at the $t$-th layer can be demonstrated as:
\begin{align}
&\{\theta_i^{m,t}|t\leq \textbf{D}_i^m\}_{m=1}^M|x_i=\textbf{LGP}^t(I_i^1,\ldots,I_i^t) \label{eq:lq} \\
    &\text{ s.t. } \Delta_i^t=\frac{1}{N^{m,t}}\sum_{m=1}^{M}{\mathds{1}(t\leq \textbf{D}_i^m)\cdot \Delta\delta_{f,i}^{m,t}}\text{ is maximized} \notag
\end{align}
where $\Delta\delta_{f,i}^{m,t}=(max(\delta_f(\theta_i^{m,t}(I_i^{m,t}))-\delta_f(I_i^{m,t}),\textbf{0}))$. $\delta_f(\cdot)$ is the feature-level quality estimation defined in Eq.(\ref{eq:feature-level-quality}). $N^{m,t}=\sum_{m=1}^M{\mathds{1}(t\leq \textbf{D}_i^m)}$. $\theta_i^{m,t} (t\leq \textbf{D}_i^m)$ is the parameter value of the $m$-th unimodal network at the $t$-th depth predicted by $\textbf{LGP}^t$. As shown in Fig. \ref{fig:lgp}, the detailed implementation of $\textbf{LGP}^t$ includes a LSTM unit $\textbf{LU}$ to remember the enhanced feature vectors $I_i^1,\ldots,I_i^t$ output by previous $t-1$ depths. As $I_i^t$ input into $\textbf{LU}$, the output modality-shared parameter information $\textbf{o}^t$ is then decoded into modality-specific parameters for the block of each unimodal network by modality-specific parameter decoders $\{\textbf{PD}^m\}_{m=1}^M$, which can be formulated as:
\begin{align}
    \theta_i^{m,t}=[W_i^{m,t},b_i^{m,t}]=\textbf{PD}^{m}(\textbf{o}^t),\text{ if }t \leq \textbf{D}_i^m.
    \label{eq:pc}
\end{align}
The pseudo-code of \textbf{\textit{LGP}} is provided in Algorithm \ref{alg:lgp}. To guide \textbf{\textit{LGP}} in predicting the optimal network parameters at each depth, in addition to maximizing $\Delta_i^t$, a sparsity loss $\mathcal{L}^s$ (Eq.(\ref{eq:L^s})) is introduced to guide the prediction of network parameters that yield the sparsest feature informative vectors $q_i^{m,t}$ (mentioned in Eq.(\ref{eq:q-enhance})), ensuring maximal removal of irrelevant and redundant information at each layer.
\begin{align}
    \mathcal{L}^{s}&=\frac{1}{N}\sum_{i=1}^N{\sum_{m=1}^M{\sum_{t=1}^{\textbf{D}_i^{m}}{||q_i^{m,t}||_1}}}. \label{eq:L^s}
\end{align}

\subsection{Learning}
During the training stage, we first optimize the noise-free prototypes via maximizing Eq.(\ref{eq:rob}). Then, the overall model is trained by minimizing the loss $\mathcal{L}$ defined as:
\begin{align}
    \mathcal{L} = \mathcal{L}^{task}-\Delta+\mathcal{L}^{s}.
\end{align}
$\mathcal{L}^{task}$ is the classification loss mentioned in Sec. \ref{sec:backbone}. $\Delta=\frac{1}{N}\sum_i{\sum_t{\Delta_i^t}}$ where $\Delta_i^t$ is defined in Eq.(\ref{eq:lq}). $\mathcal{L}^{s}$ is the sparsity loss formulated in Eq.(\ref{eq:L^s}).

\section{Experiments}
In this section, the effectiveness of the proposed Multi-QuAD is evaluated compared with 7 state-of-the-art multimodal reliable classification methods and multimodal dynamic network on four benchmarks. In the following section, we will first introduce the experimental settings in Sec. \ref{sec:exp-set}. Then, the comparison with SOTA methods, ablation study, discussion on three key component of Multi-QuAD (\textbf{\textit{NFCE}}, \textbf{\textit{GCND}}, and \textbf{\textit{LGP}}), analysis on learned features and computation effectiveness will be presented respectively in Sec. \ref{sec:comp-sota}, \ref{sec:ablation}, \ref{sec:dis-nfce}, \ref{sec:gcnd}, \ref{sec:dis-lgp}, \ref{sec:dis-fea}, and \ref{sec:dis-ce}.


\subsection{Experimental Setups} \label{sec:exp-set}
\subsubsection{Datasets} \label{sec:dataset}
Experiments are conducted on four commonly used multimodal datasets in previous methods \cite{han2020trusted,geng2021uncertainty,han2022multimodal,zhang2023provable,zheng2023multi,zhou2023calm,pmlr-v235-cao24c}. (1) \textbf{BRCA} is a dataset for classifying breast invasive carcinoma into 5 PAM50 subtypes, containing 875 samples with features from three modalities: mRNA expression, DNA methylation, and miRNA expression. It is available from The Cancer Genome Atlas (TCGA) \footnote{\url{https://www.cancer.gov/aboutnci/organization/ccg/research/structuralgenomics/tcga}}. (2) \textbf{ROSMAP} contains samples from Alzheimer's patients (182 samples) and normal control subjects (169 samples), including the same modalities as \textbf{BRCA}. \cite{a2012overview,de2018multi} (3) \textbf{CUB} is an image-text dataset with 11,788 samples comprising 200 categories of birds. \cite{wah2011caltech} (4) \textbf{UPMC FOOD101} comprises 90,704 food images from 101 categories and corresponding textual descriptions \cite{wang2015recipe}.

\subsubsection{Compared methods} \label{sec:comp-method}
To demonstrate the improvement in multimodal classification reliability achieved by Multi-QuAD, several representative methods in this field are introduced, including multimodal dynamics (\textbf{MD}) \cite{han2022multimodal}, multi-level confidence learning (\textbf{MLCLNet}) \cite{zheng2023multi}, dynamic poly-attention network (\textbf{DPNET}) \cite{zou2023dpnet}, enhanced encoding and confidence evaluating framework (\textbf{CALM}) \cite{zhou2023calm}, quality-aware multimodal fusion (\textbf{QMF}) \cite{zhang2023provable}, multi-omics data classification network via global and cross-modal feature aggregation (\textbf{GCFANet}) \cite{zheng2024global}, and predictive dynamic fusion (\textbf{PDF}) \cite{pmlr-v235-cao24c}. To demonstrate the advantages of our method over other multimodal dynamic neural networks, \textbf{DynMM} \cite{xue2023dynamic} is also included in the experiments. Other dynamic neural networks are excluded from the experiments as they are not applicable to multimodal scenarios.

\subsubsection{Evaluation metrics} \label{sec:eval-metrics}
For multi-class classification tasks, three metrics, including accuracy (ACC), average F1 score weighted by support (WeightedF1), and macro-averaged F1 score (MacroF1), are employed to evaluate the performance of different methods. For binary classification tasks, ACC, F1 score (F1), and area under the receiver operating characteristic curve (AUC) of different methods are compared.

\begin{figure}[t]
\centering
\includegraphics[width=0.95\columnwidth]{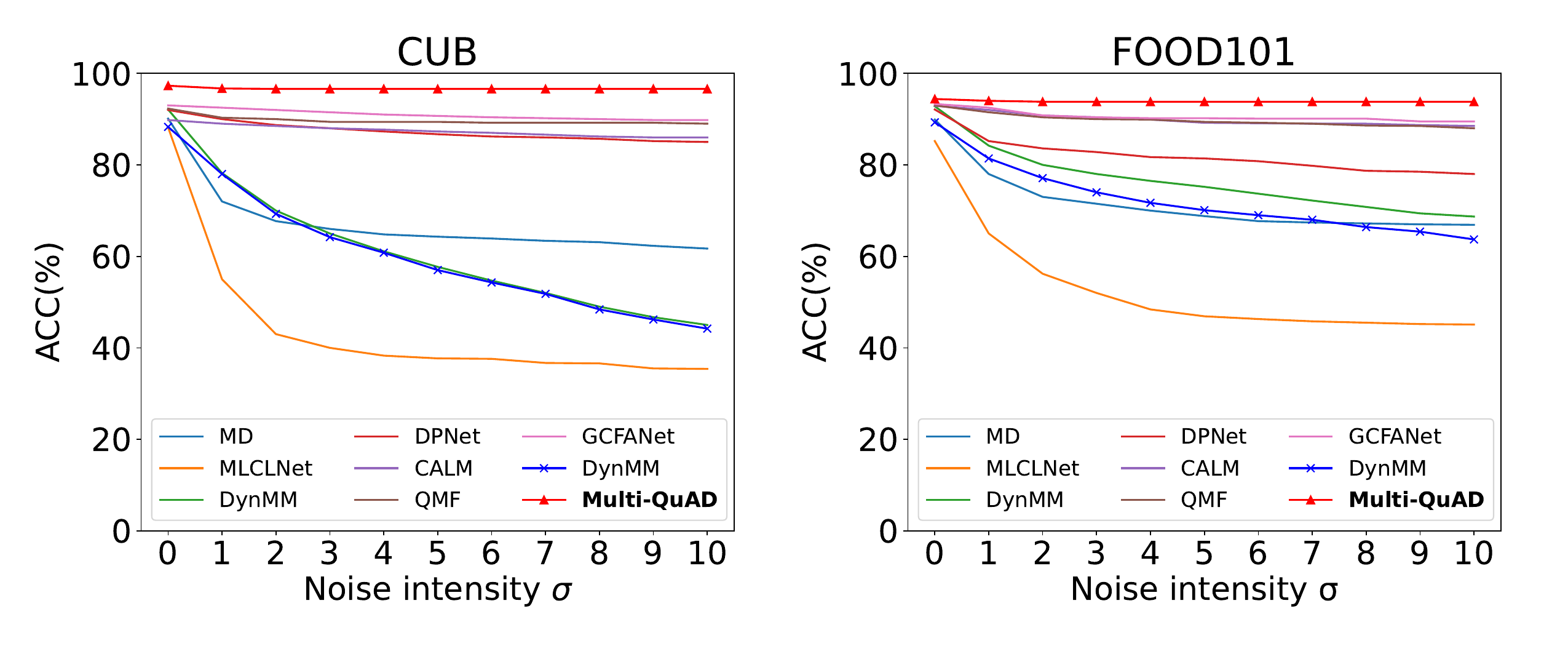} 
\caption{Comparison of classification accuracy of different models under Gaussian noise with different intensities $\sigma$.}
\label{fig:noise}
\end{figure}

\begin{figure}[t]
  \centering
  \includegraphics[width=0.95\linewidth]{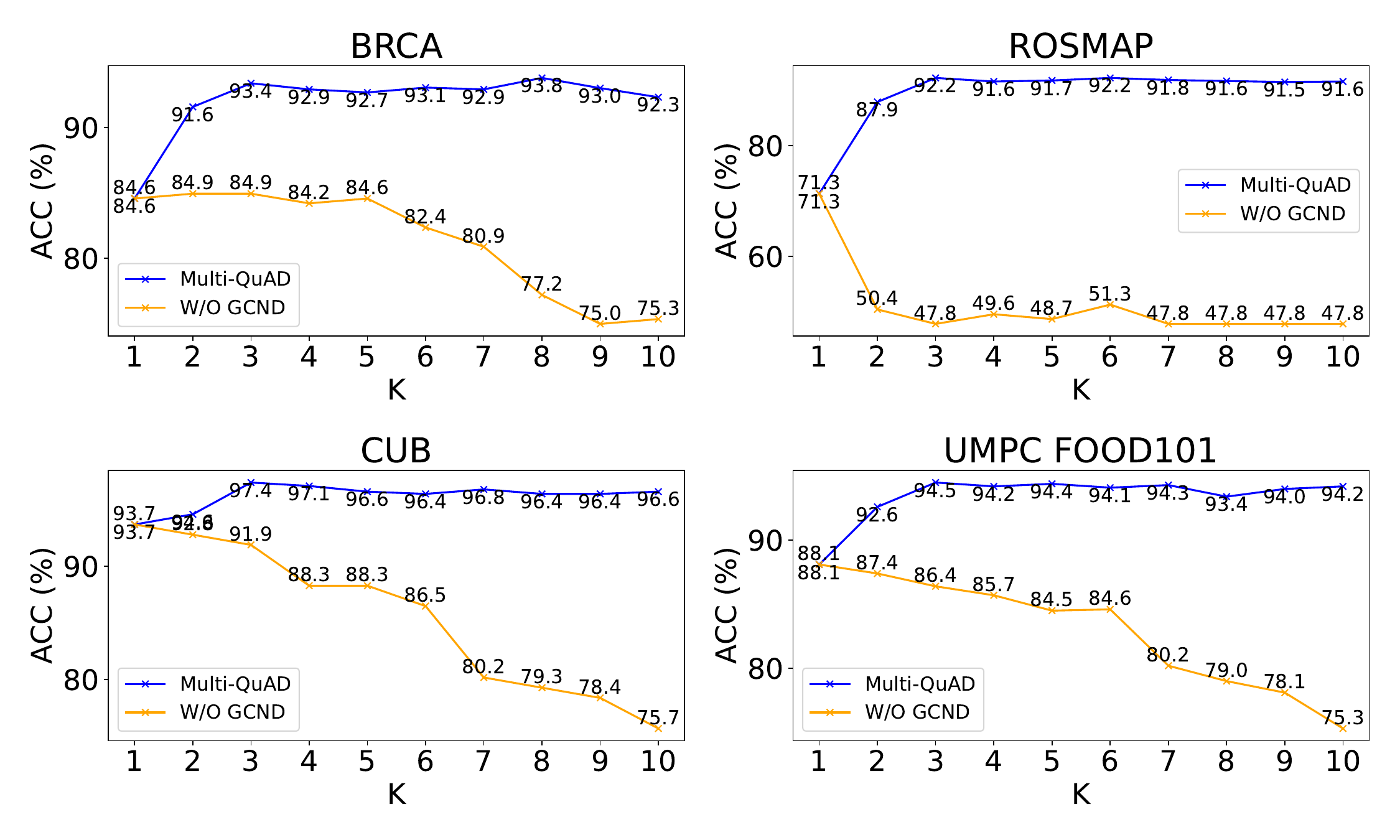}
  \caption{Ablation study to demonstrate the effectiveness of \textbf{\textit{GCND}}.}
  \label{fig_ablation_dd}
\end{figure}

\subsubsection{Implementation details} \label{sec:imple-detail}

Noise is added to simulate low-quality data in the experiments, with $\sigma$ representing the noise intensity. We run experiments 20 times to report the mean and standard deviation, following previous work \cite{han2022multimodal}. The Adam optimizer and the step learning rate scheduler with an initial learning rate of 1e-4 are employed in the training. \textbf{More details are provided in the Supplemental Materials.}


\subsection{Comparison With State-of-The-Art Methods}\label{sec:comp-sota}

Following previous works \cite{han2020trusted,ma2021trustworthy,zhang2023provable,zhou2023calm}, we compare the classification performance of different methods under noise-free and Gaussian noise-added settings, as shown in Table \ref{sota table1} and Fig. \ref{fig:noise}, respectively.

\textbf{(1) Comparison with state-of-the-art reliable multimodal classification methods.} Results in Table \ref{sota table1} show that the proposed Multi-QuAD outperforms SOTA reliable multimodal classification methods such as \textbf{CALM}, \textbf{GCFANet}, and \textbf{PDF} by a substantial margin under the noise-free setting on four datasets. On BRCA, ROSMAP, CUB, and FOOD101, Multi-QuAD achieves ACC gains of 4.8\%, 5.9\%, 4.4\%, and 1.2\% over the leading approach, respectively. The possible reason is that Multi-QuAD's dynamic network capability enables it to learn distinct mapping relationships for different samples, thereby enhancing the precision of the predictions. Under noise setting (Fig. \ref{fig:noise}), Multi-QuAD exhibits more stable performance compared to SOTA methods, with a larger ACC improvement than under noise-free conditions. This indicates that the reliable quality estimation method, the sample-specific network depth, and sample-specific parameters endow Multi-QuAD with a more robust performance.

\textbf{(2) Comparison with multimodal dynamic neural network.} Multi-QuAD outperforms \textbf{DynMM} across four datasets under the noise-free setting as shown in Table \ref{sota table1}. This is primarily because Multi-QuAD's dynamic parameter mechanism provides stronger representation capability. While under Gaussian noise (Fig. \ref{fig:noise}), Multi-QuAD's more reliable quality estimation method and dynamic depth mechanism make its performance much more stable than \textbf{DynMM}. \textbf{DynMM} suffers significant performance degradation as noise intensity $\sigma$ increases. This is likely because it adjusts excessively deep network layers for hard samples, leading to overfitting and unreliable results.

\begin{table}[]
\caption{Ablation study on four datasets to demonstrate the effectiveness of \textbf{\textit{LGP}}, where the best results are in bold.}
\begin{center}
\begin{tabular}{c|c|c|c|c}
\hline
Dataset               & Method & $\sigma=0$ & $\sigma=5$ & $\sigma=10$\\ \hline
\multirow{2}{*}{BRCA}   & W/O LGP     & 86.2 & 83.4 & 80.2  \\
                    & Multi-QuAD      & \textbf{93.4} & \textbf{92.4} & \textbf{91.5}\\ 
\hline

\multirow{2}{*}{ROSMAP}  & W/O LGP     & 85.4 & 82.6 & 80.4   \\
                      & Multi-QuAD      & \textbf{92.2} & \textbf{91.6} & \textbf{90.8} \\ 
\hline
\multirow{2}{*}{CUB} & W/O LGP     & 90.3 & 86.2 & 83.6 \\
                      & Multi-QuAD     & \textbf{97.4} & \textbf{96.7}& \textbf{95.5} \\ 
                      \hline

\multirow{2}{*}{FOOD101}  & W/O LGP     & 90.2 & 87.7 & 85.1   \\
                      & Multi-QuAD      & \textbf{94.5} & \textbf{93.8} & \textbf{93.0}  \\ 
                      \hline
\end{tabular}
\end{center}
\label{tab_ablation_dp}
\end{table}

\subsection{Ablation study} \label{sec:ablation}
Ablation studies are conducted on all four datasets to demonstrate the effectiveness of the Global Confidence Normalized Depth (\textbf{\textit{GCND}}) and Layer-wise Greedy Parameter (\textbf{\textit{LGP}}) mechanisms in enhancing the reliability of Multi-QuAD. To this end, two degraded model versions, ``W/O GCND'' and  ``W/O LGP'', are introduced. In ``W/O GCND'', the depth of each unimodal model is fixed to the hyperparameter $K$. In ``W/O LGP'', the parameters of all network blocks are determined during training instead of being dynamically predicted during inference. \textbf{(i) Effectiveness of \textit{GCND}.} Fig. \ref{fig_ablation_dd} shows the classification accuracy of ``W/O GCND'' and Multi-QuAD on four datasets. When $K=1$, the two network structures are identical, resulting in the same accuracy. When $K\geq 2$, Multi-QuAD's unimodal networks can dynamically adjust their depth between 1 and $K$. In contrast, ``W/O GCND'' maintains a fixed depth of $K$, which becomes too deep for a significant portion of easy samples in the dataset, causing overfitting and a notable performance drop. As $K$ increases, the issues with ``W/O GCND'' become more severe, resulting in a continuous decline in accuracy. \textbf{(ii) Effectiveness of \textit{LGP}.} Table \ref{tab_ablation_dp} compares the classification accuracy of ``W/O LGP'' and Multi-QuAD on four datasets under different noise intensities. Multi-QuAD outperforms ``W/O LGP'' across different noise levels, and its performance degradation as noise intensity increases is significantly less severe than ``W/O LGP''. This demonstrates that \textbf{\textit{LGP}} contributes to both higher classification performance and improved robustness.


\begin{figure}[t]
  \centering
  \includegraphics[width=0.97\linewidth]{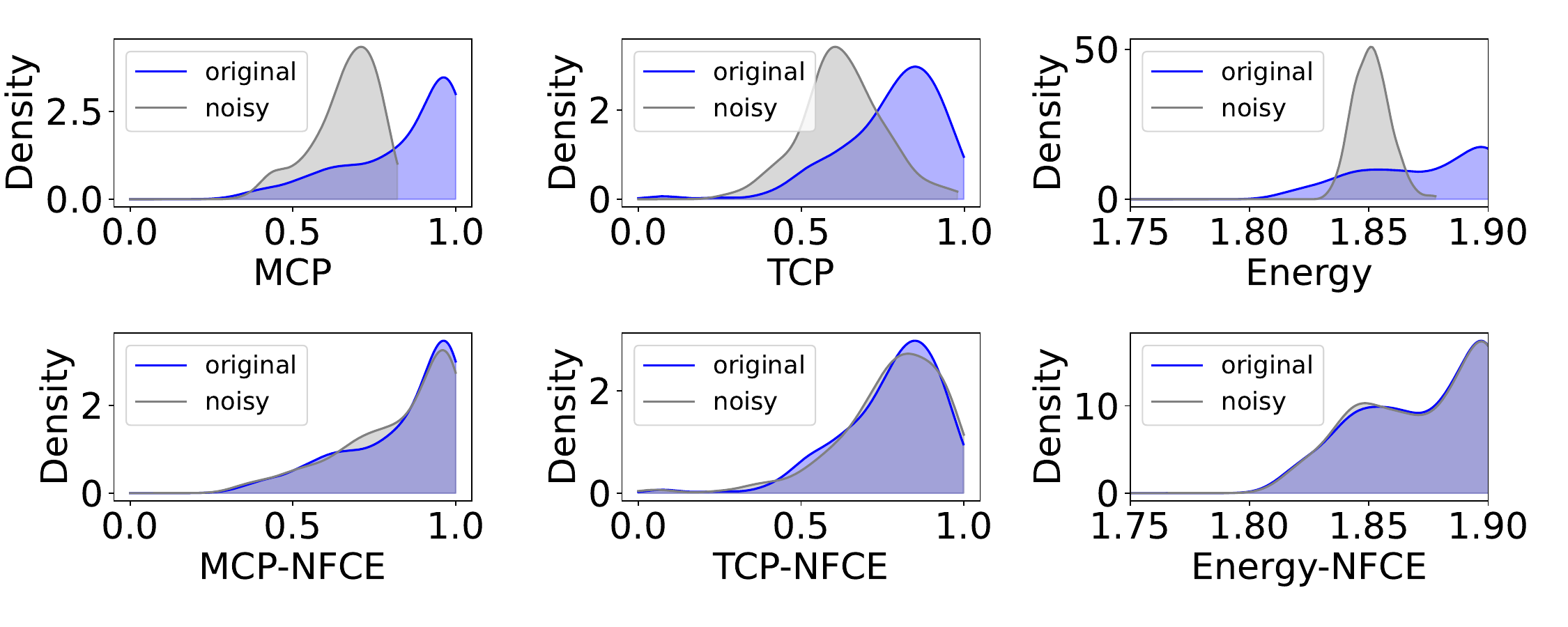}
  \caption{Comparison of the sensitivity of some commonly used traditional confidence assessment methods and their “-NFCE” counterparts to training noise on the FOOD101 dataset.}
  \label{fig_compare_conf}
\end{figure}

\begin{figure}[t]
    \centering
    \subfloat[Results on the BRCA dataset.]{
        \includegraphics[width=\columnwidth]{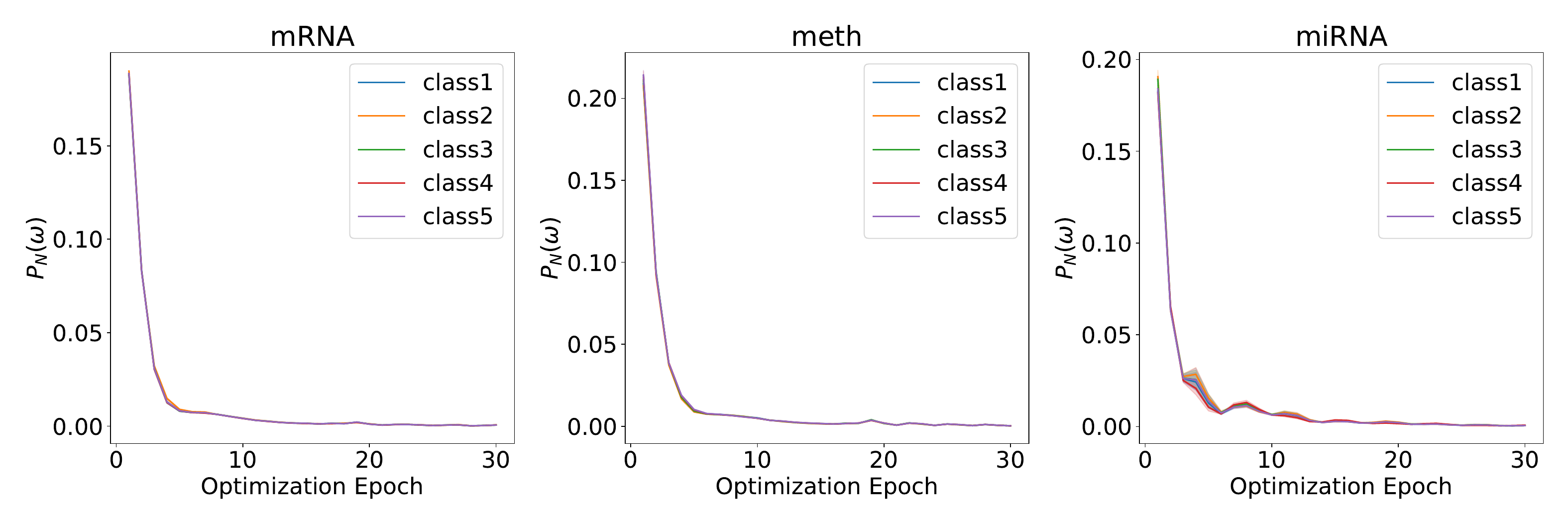}
        \label{fig:noise-free-brca}
    }
    \hfill
    \subfloat[Results on the UPMC FOOD101 dataset.]{
        \includegraphics[width=0.95\columnwidth]{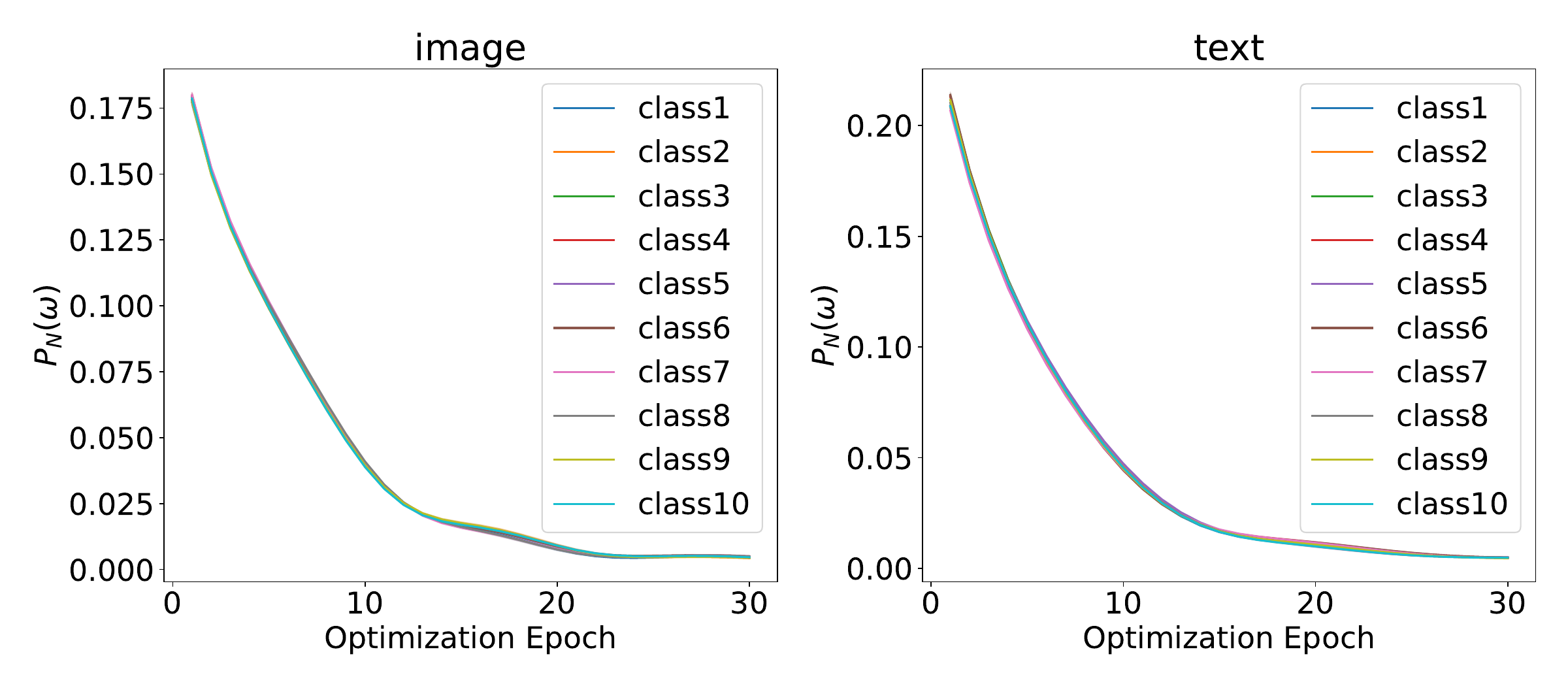}
        \label{fig:noise-free-food}
    }
    \caption{The probability of noise-free prototypes in the noise distribution $P_{\mathcal{N}}(\omega)$ at different optimization epochs.}
    \label{fig:main}
\end{figure}

\begin{table}[]
\caption{Results of Multi-QuAD on the FOOD101 dataset using different confidence estimation methods.}
\begin{center}
\begin{tabular}{c|c|c|c}
\hline
Method & $\sigma=0$ & $\sigma=1$ & $\sigma=5$ \\ \hline
Multi-QuAD (Cls) & 93.5 & 91.7 & 90.4 \\ \hline
Multi-QuAD (NFCE) & \textbf{94.5} & \textbf{94.0} & \textbf{93.8}  \\
\hline
\end{tabular}
\end{center}
\label{tab_conf_compare}
\end{table}




\subsection{Discussion on \textbf{NFCE}} \label{sec:dis-nfce}
\subsubsection{The Reliability of \textbf{NFCE}}
We conduct experiments on commonly used traditional confidence estimation methods, including MCP, TCP \cite{han2022multimodal}, and Energy \cite{liu2020energy}, all of which assess confidence based on class probabilities obtained by network or classifier. To demonstrate that the proposed noise-free prototype confidence estimation can enhance the reliability of these methods, we replaced their class probability computing approach with the method described in Eq.(\ref{eq:conf-m}), resulting in ``MCP-NFCE'', ``TCP-NFCE'', and ``Energy-NFCE'', respectively. Each method is trained separately on the original and noisy training set with Gaussian noise ($\sigma=5$). Fig. \ref{fig_compare_conf} reports the confidence estimation of the same clean testing samples provided by two trained versions of each method. The testing results obtained by the same method after training on original and noisy data are expected to be nearly identical since the confidence evaluation is performed on the same test samples. However, the results of the three traditional methods exhibit shifts. Notably, the testing confidence corresponding to the noisy training set is significantly lower than that of the original. The reason is that the classifiers or network structures in these methods overfit the noise distribution, resulting in reduced classification confidence for clean test samples. In contrast, the three ``-NFCE'' counterparts maintain stable testing confidence for different training noises. Thanks to the noise-free prototypes and the classifier-free design, the confidence of samples can be evaluated more reliably. 

Furthermore, we substitute the method of Eq.(\ref{eq:conf-m}) in Multi-QuAD with classifier and yield ``Multi-QuAD (Cls)''. Table \ref{tab_conf_compare} compares the classification accuracy of ``Multi-QuAD (Cls)'' and original Multi-QuAD with \textbf{\textit{NFCE}} under different noise intensities. The results indicate that \textbf{\textit{NFCE}} provides Multi-QuAD with greater reliability than classifier-based. 

\begin{table}[t]
\caption{Comparison of classification accuracy under different noise intensities for Multi-QuAD and its ``depth-plus-one version'' and ``depth-minus-one version.''}
\begin{center}
\begin{tabular}{c|c|c|c|c}
\hline
Dataset               & Method & $\sigma=0$ & $\sigma=5$ & $\sigma=10$\\ \hline
\multirow{3}{*}{BRCA}   & Multi-QuAD(-1)     & 92.8 & 91.4 & 90.2  \\
                        & Multi-QuAD(+1)     & 92.9 & 91.6 & 90.1  \\
                    & Multi-QuAD      & \textbf{93.7} & \textbf{92.4} & \textbf{91.5}\\ 
\hline

\multirow{3}{*}{ROSMAP}  & Multi-QuAD(-1)     & 91.6 & 90.8 & 90.1  \\
                        & Multi-QuAD(+1)     & 91.5 & 90.8 & 90.2  \\
                      & Multi-QuAD      & \textbf{92.1} & \textbf{91.6} & \textbf{90.8} \\ 
\hline
\multirow{3}{*}{CUB} & Multi-QuAD(-1)     & 96.4 & 95.7 & 95.0  \\
                        & Multi-QuAD(+1)     & 96.5 & 95.7 & 94.9  \\
                      & Multi-QuAD     & \textbf{97.3} & \textbf{96.7}& \textbf{95.5} \\ 
                      \hline

\multirow{3}{*}{FOOD101}  & Multi-QuAD(-1)     & 93.5 & 92.9 & 92.3  \\
                        & Multi-QuAD(+1)     & 93.5 & 92.8& 92.2  \\
                      & Multi-QuAD      & \textbf{94.4} & \textbf{93.7} & \textbf{93.0}  \\ 
                      \hline
\end{tabular}
\end{center}
\label{tab:gcnd_opt_depth}
\end{table}

\subsubsection{Noise Removal of the Noise-Free Prototypes (The Effectiveness of $\mathcal{L}^{rob}$)} To verify that the noise is effectively removed in prototypes during optimization via $\mathcal{L}^{rob}$ formulated in Eq.(\ref{eq:rob}), noise with $\sigma=5$ is added to all modalities of the datasets. At each optimization epoch, we report the probability of the prototype corresponding to each class and modality within the noise distribution: $P_{\mathcal{N}}(\omega)=\{\{P_{\mathcal{N}}(\omega^{c,m})\}_{c=1}^C\}_{m=1}^M$. $P_{\mathcal{N}}$ is the probability density function of the added noise. The results in Fig. \ref{fig:noise-free-brca} and \ref{fig:noise-free-food} shows that as the optimization progresses, the probabilities $P_{\mathcal{N}}(\omega)$ quickly converge to nearly 0 after a few epochs, indicating that the noise in the prototypes is effectively removed using $\mathcal{L}^{rob}$.

\subsubsection{Multi-level quality estimation} 
\begin{figure}[t]
    \centering
    \subfloat[The quality distributions of clean and noisy halves of the data in different modalities on the UPMC FOOD101 dataset.]{
        \includegraphics[width=0.9\columnwidth]{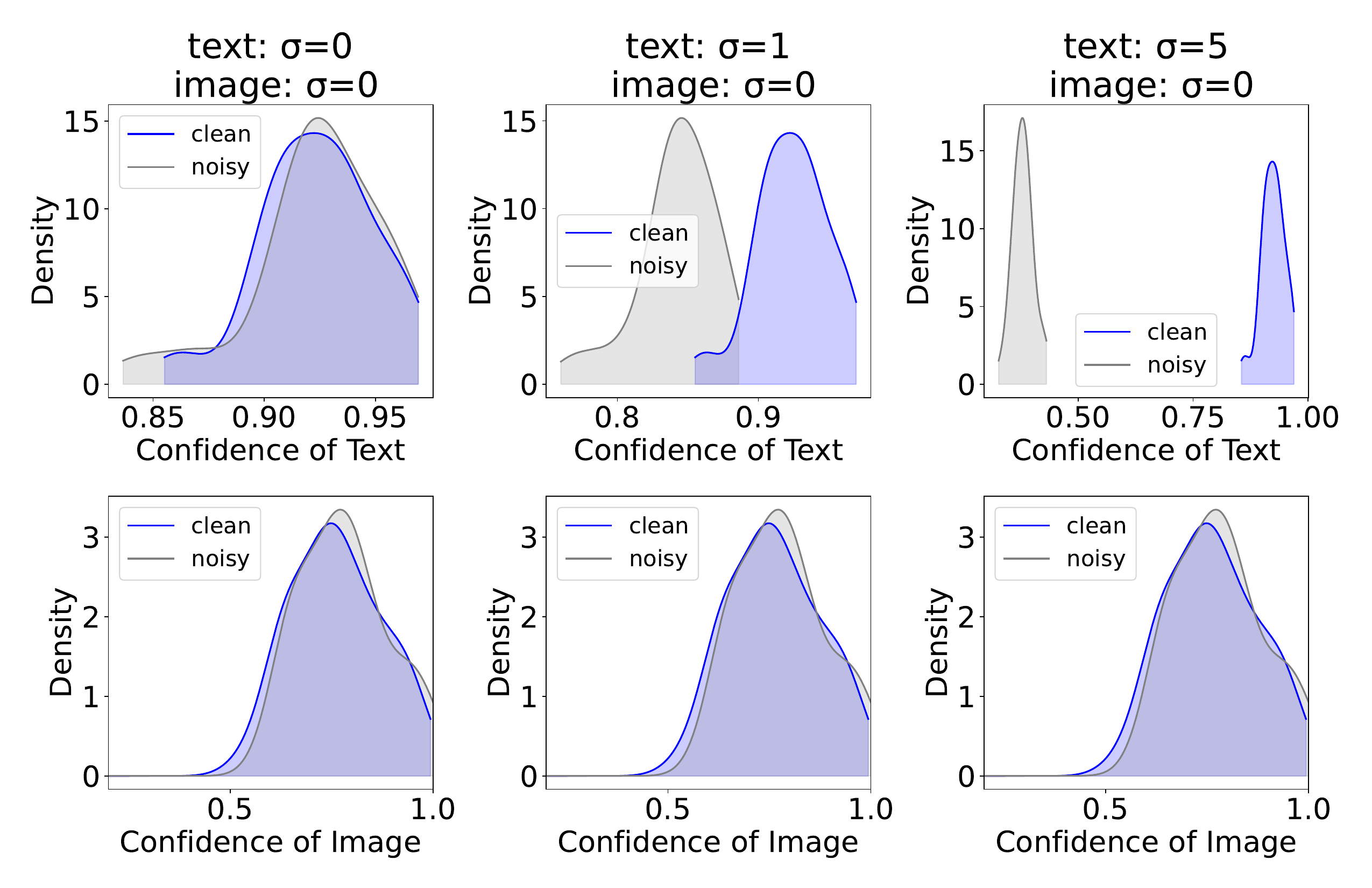}
        \label{fig:modality-lv-q}
    }
    \hfill
    \subfloat[The feature-level quality of the noisy data on the UPMC FOOD101 dataset under different noise intensities.]{
        \includegraphics[width=\columnwidth]{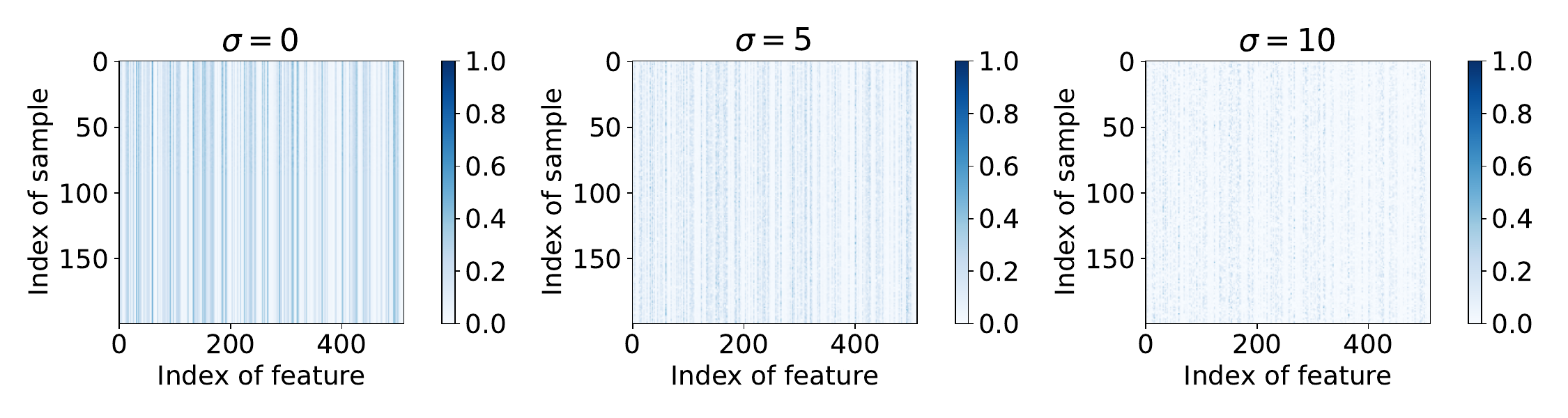}
        \label{fig:feature-lv-q}
    }
    \caption{Results of the modality-level and feature-level quality estimated by \textbf{\textit{NFCE}}.}
    \label{fig:quality-est}
\end{figure}

Extensive experiments are conducted to verify \textbf{\textit{NFCE}}'s capability to provide modality-level and feature-level quality estimation. \textbf{(i) Modality level}. As shown in Fig. \ref{fig:modality-lv-q}, noise is added to the text modality of half the UPMC FOOD101 dataset samples, and the quality distributions of both modalities calculated using $\delta_v$ (Eq.(\ref{eq:modality-level-quality})) in these two halves are reported. When the noise intensity $\sigma=0$, the quality distributions of the two halves of the text modality samples are nearly identical. As noise intensity in text modality increases, the quality distribution of the noisy text data becomes increasingly skewed to the left, while the unaffected image modality maintains consistent quality between the two halves of the data. This indicates that \textbf{\textit{NFCE}} can efficiently sense changes in modality quality. \textbf{(ii) Feature level}. Fig. \ref{fig:feature-lv-q} visualizes the quality of the noisy half of text modality, with each element in each row of the heatmap representing the quality of each feature in a sample's feature vector. The distinct values in each column of the heatmap indicate that \textbf{\textit{NFCE}} can perceive the varying qualities of different feature values among samples calculated using $\delta_f$ (Eq.(\ref{eq:feature-level-quality})). As the noise intensity increases, the feature-level quality of the samples decreases, and the overall color of the heatmap becomes lighter. This indicates that \textbf{\textit{NFCE}} can perceive the changes in feature-level quality brought about by noise intensity.


\subsection{Discussion on the \textbf{GCND}}\label{sec:gcnd}
\subsubsection{\textbf{GCND} Adjusts Optimal Network Depth For Samples} Two modified versions of Multi-QuAD terms ``Multi-QuAD(-1)'' and ``Multi-QuAD(+1)'' are introduced to verify whether \textbf{\textit{GCND}} adjusts the optimal depth for samples. ``Multi-QuAD(-1)'' employs network depths that are one layer shallower than Multi-QuAD. Specifically, the network depths of ``Multi-QuAD(-1)'' for sample $x_i$ is $\textbf{D}_i^{\prime}=\{max(\textbf{D}_i^m-1,1)\}_{m=1}^M$, where $\textbf{D}_i=\{\textbf{D}_i^m\}_{m=1}^M$ is the depth adjusted by Multi-QuAD. ``Multi-QuAD(+1)'' employs network depths that are one layer deeper than Multi-QuAD, i.e., $\textbf{D}_i^{\prime}=\{\textbf{D}_i^m+1\}_{m=1}^M$. We add noise with different intensities $\sigma$ and compare the classification accuracy of ``Multi-QuAD(-1)'', ``Multi-QuAD(+1)'', and Multi-QuAD on all datasets. Results in Table \ref{tab:gcnd_opt_depth} demonstrate that both ``Multi-QuAD(-1)'' and ``Multi-QuAD(+1)'' consistently underperform compared to Multi-QuAD across all four datasets and various noise intensities. This indicates that Multi-QuAD adjusts the most reliable depths for samples.


\begin{figure}[t]
\centering
\includegraphics[width=\columnwidth]{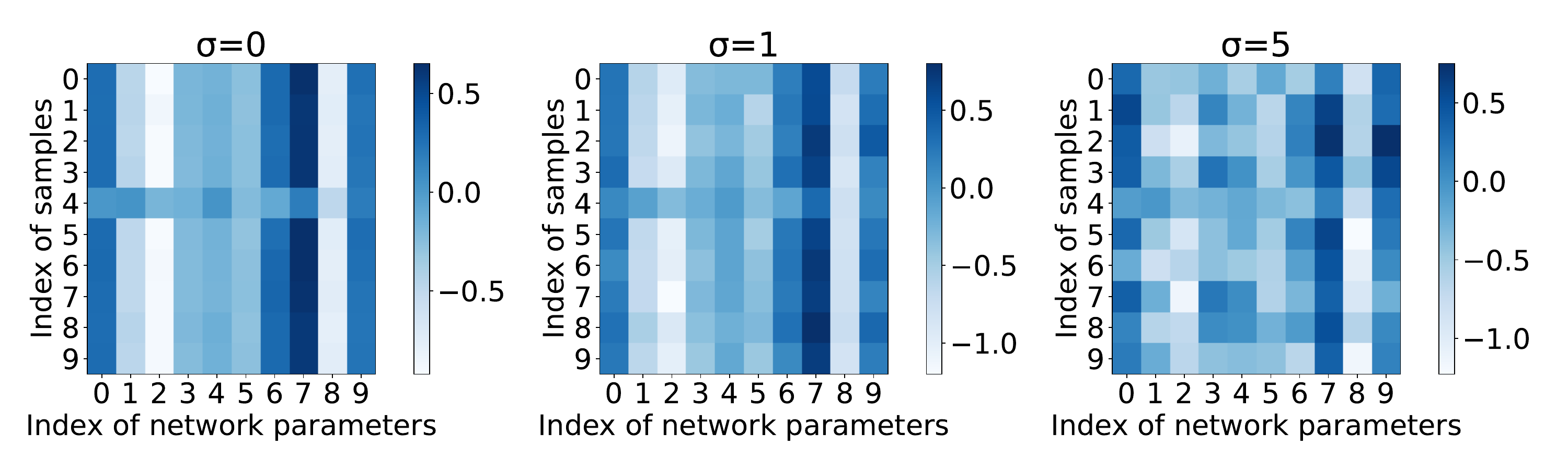} 
\caption{Visualization of the parameter values of different samples on UMPC FOOD101 dataset under different noise intensities. For clarity, the first 10 elements of the parameters are shown.}
\label{fig:vis-dyparam}
\end{figure}

\begin{table}[]
\caption{Classification accuracy of \textbf{ICD} and \textbf{\textit{GCND}} on four datasets under different noise intensities.}
\begin{center}
\begin{tabular}{c|c|c|c|c}
\hline
Dataset               & Method & $\sigma=0$ & $\sigma=5$ & $\sigma=10$\\ \hline
\multirow{2}{*}{BRCA}   & ICD     & 92.0 & 89.4 & 87.2  \\
                    & GCND      & \textbf{93.4} & \textbf{92.4} & \textbf{91.5}\\ 
\hline

\multirow{2}{*}{ROSMAP}  & ICD     & 90.4 & 88.2 & 86.5  \\
                      & GCND      & \textbf{92.2} & \textbf{91.6} & \textbf{90.8} \\ 
\hline
\multirow{2}{*}{CUB} & ICD     & 94.6 & 92.1 & 89.4  \\
                      & GCND     & \textbf{97.4} & \textbf{96.7}& \textbf{95.5} \\ 
                      \hline

\multirow{2}{*}{FOOD101}  & ICD     & 92.1 & 90.5 & 88.6  \\
                      & GCND      & \textbf{94.5} & \textbf{93.8} & \textbf{93.0}  \\ 
                      \hline
\end{tabular}
\end{center}
\label{tab:gcnd-comp-trad}
\end{table}

\subsubsection{\textit{\textbf{GCND}} Exhibits Stronger Reliability Than Conventional Dynamic Depth Method} Existing dynamic depth methods typically rely on intermediate modules to determine network depth, such as the gating network in DynMM \cite{xue2023dynamic} and the intermediate classifier in MSDNet \cite{huang2017multi}. The design of these intermediate modules prevents the network depth from being determined before sample input, making the methods prone to excessive depth for difficult samples during inference and thus affecting model reliability. In contrast, \textbf{\textit{GCND}} enhances the reliability of dynamic depth by mitigating the adverse effects of extreme modality samples, as introduced in Sec. \ref{sec:dydepth}. To validate this, we implement the intermediate-classifier-based dynamic depth termed ``\textbf{ICD}'' following that in MSDNet, and replace \textbf{\textit{GCND}} in Multi-QuAD with it. The classification accuracy of the two implementations of dynamic depth under different noise intensities is illustrated in Table \ref{tab:gcnd-comp-trad}. It is evident that \textbf{\textit{GCND}} enables Multi-QuAD to achieve higher and more stable classification accuracy than \textbf{ICD} under varying noise levels, demonstrating its stronger reliability.

\subsubsection{Analysis of Hyperparameter $K$}\label{sec:hyper-param-analy} There is one hyperparameter $K$ in \textit{\textbf{GCND}} (mentioned in Sec. \ref{sec:dydepth}). The sensitivity analysis is conducted by varying $K$ in the range $[1,10]$. The result in the purple curve of Fig. \ref{fig_ablation_dd} shows that Multi-QuAD is relatively insensitive to $K$ in all datasets as $K$ is greater than or equal to 3. \textbf{Due to space limitation, more discussions on \textit{GCND} can be found in the Supplemental Materials.}

\begin{figure}[t]
\centering
\includegraphics[width=\columnwidth]{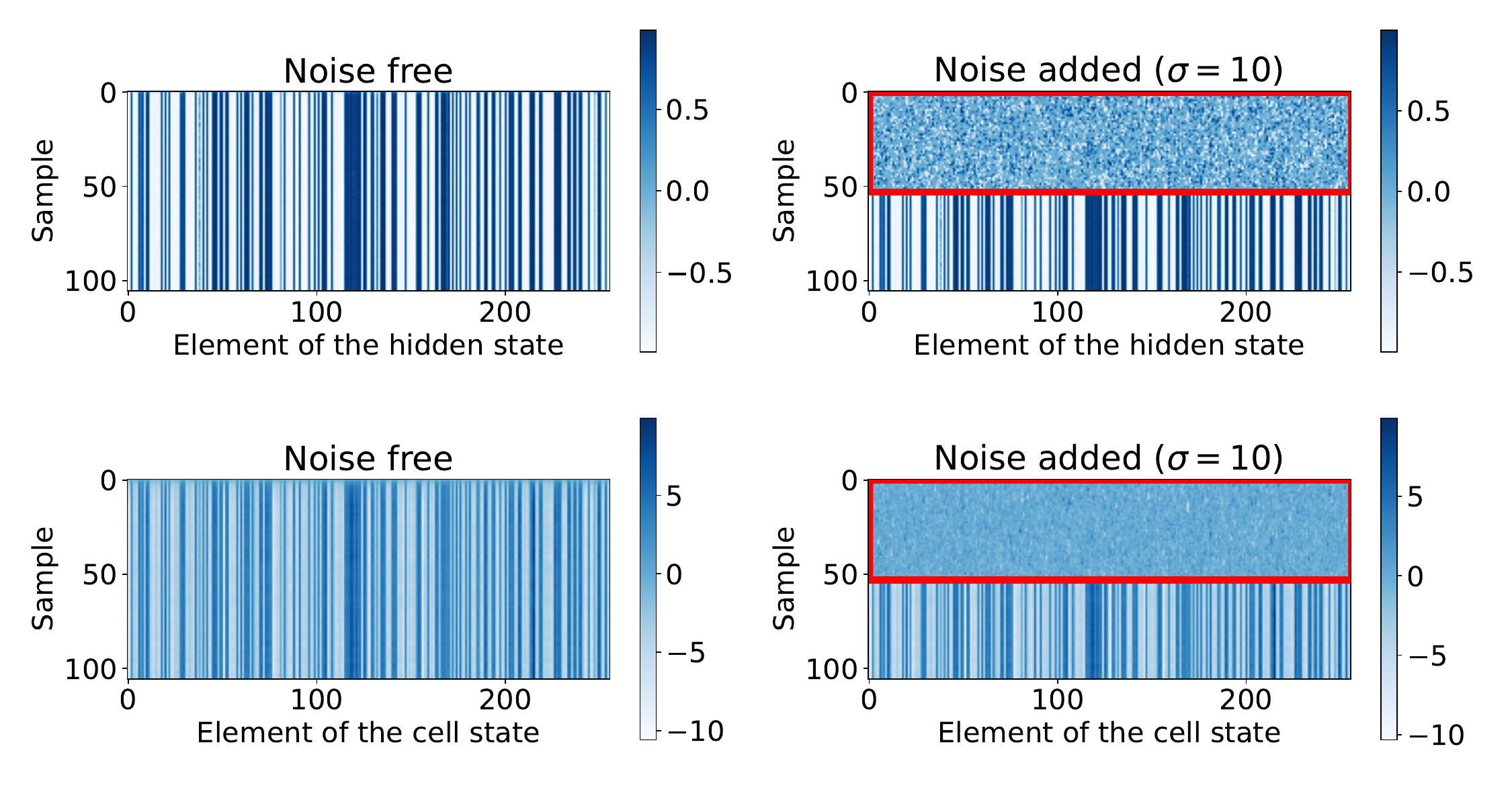} 
\caption{Visualization of the element values of the hidden state and cell state of the LSTM Unit at the first layer. The result on UPMC FOOD101 dataset is shown, the noisy half of the samples are marked with red box.}
\label{fig:vis-lu}
\end{figure}

\begin{figure}[t]
    \centering
    \subfloat[The results on the BRCA dataset.]{
        \includegraphics[width=0.9\columnwidth]{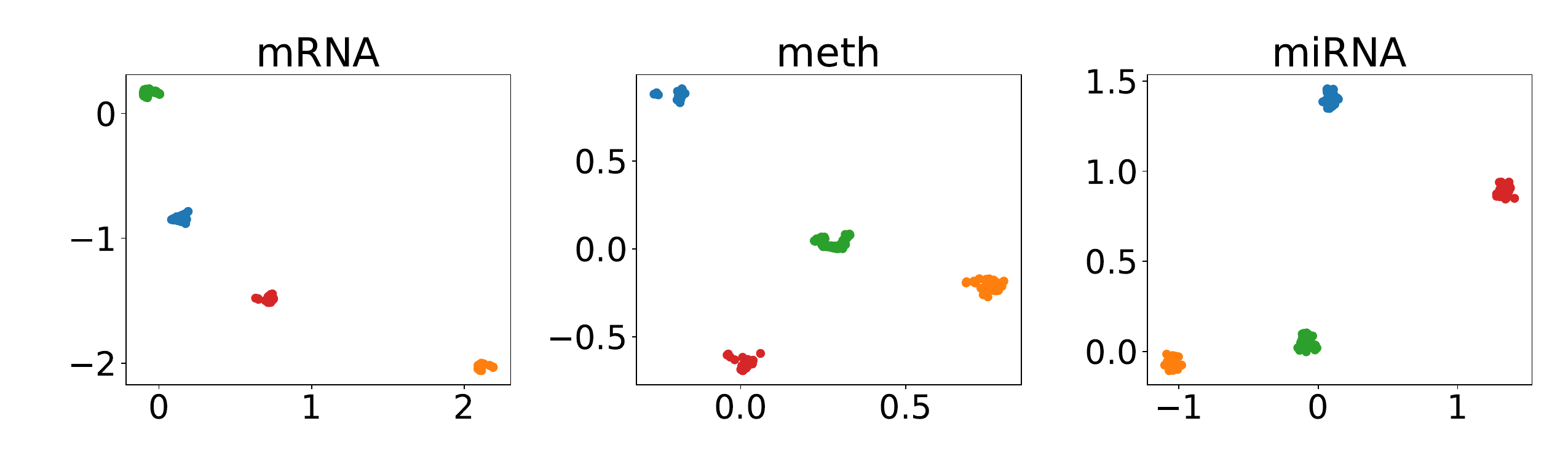}
        \label{fig:tsne-brca}
    }
    \hfill
    \subfloat[The results on the ROSMAP dataset.]{
        \includegraphics[width=0.9\columnwidth]{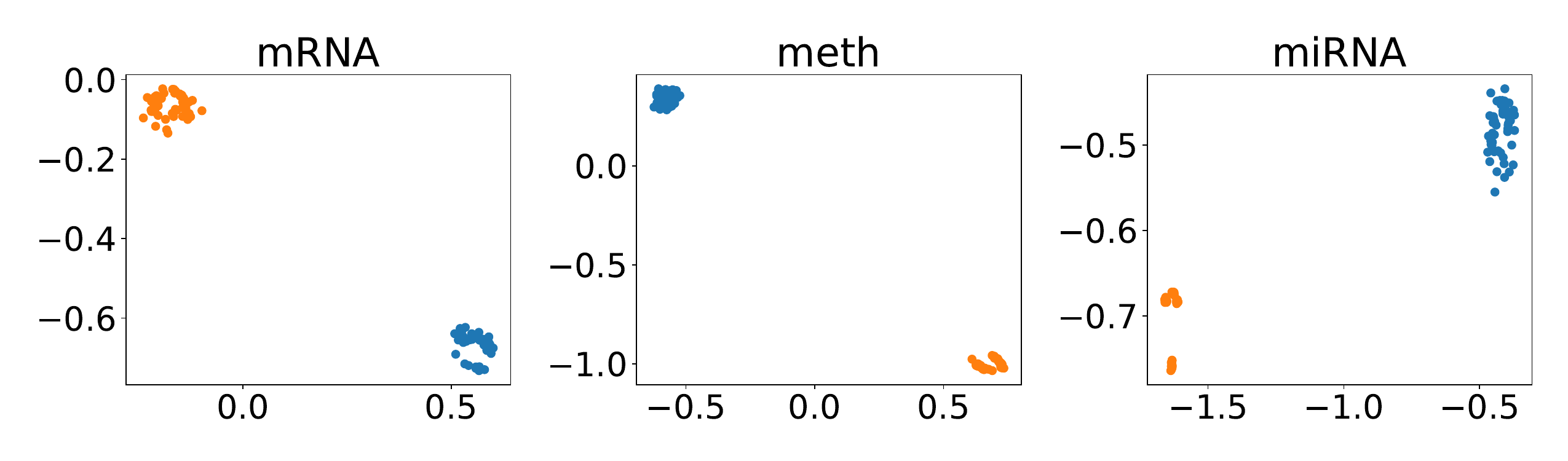}
        \label{fig:tsne-rosmap}
    }
    \caption{The t-SNE visualization results of the output feature vectors of different modalities enhanced by Multi-QuAD on the BRCA and ROSMAP datasets. Clusters of different colors represent the features of samples from different classes.}
    \label{fig:tsne}
\end{figure}

\subsection{Discussion on the \textbf{LGP}} \label{sec:dis-lgp}
\subsubsection{Parameters Adaptability to Feature-Level Quality} To demonstrate the adaptability of parameters provided by \textit{\textbf{LGP}} under feature-level quality variation, we visualize the parameters from the same quality-enhancing network block of Multi-QuAD for different samples under noise. Fig. \ref{fig:vis-dyparam} shows the parameter values of 10 randomly chosen samples of the noisy text modality on the UPMC FOOD101 dataset. By comparing the differences between the rows of the heatmaps, one can observe that \textbf{\textit{LGP}} predicts fairly consistent yet distinct network parameters for clean samples ($\sigma=0$). However, as the intensity of noise increases, introducing more chaotic features into the feature vectors, the network parameters predicted by \textbf{\textit{LGP}} for different samples become increasingly varied. This demonstrates that \textbf{\textit{LGP}} can perceive the quality variations at the feature level within the feature vectors of each sample and dynamically adjust the parameters.

\subsubsection{Effectiveness of the LSTM Unit (\textbf{LU})}
To further illustrate the mechanism of \textbf{\textit{LGP}}, we visualize the hidden state and cell state of the LSTM Unit (\textbf{LU}) under noise. Fig. \ref{fig:vis-lu} shows the results at the first layer on the UPMC FOOD101 dataset. Under noise-free conditions, the hidden state and cell state across samples are rather consistent. After the noise ($\sigma=10$) is added to half of the samples, the feature-level quality of the affected half of the samples becomes irregular and varied, and the corresponding half of the hidden state and cell state (marked with red boxes) also exhibit cross-sample variation. This demonstrates that \textbf{LU} can perceive changes in feature values across samples, thereby ensuring reliable parameter prediction. \textbf{Due to space limitation, more discussions on \textit{LGP} can be found in the Supplemental Materials.}

\subsection{Informative features that Multi-QuAD learns} \label{sec:dis-fea}
The reliability of Multi-QuAD is further demonstrated by analyzing the informative features it retains following previous works \cite{han2022multimodal,zou2023dpnet}. Table \ref{tab_fea_learn} shows the top 5 informative biomarkers of the three modalities on the BRCA dataset retained by the feature informativeness masks $q_i^{m,t}$ learned in Eq.(\ref{eq:q-enhance}) by Multi-QuAD, which align with the findings of existing biomedical research. For example, low expression of ZNF671 \cite{zhang2019epigenetic}, miR-378 \cite{arabkari2023mirna} are both associated with poor prognosis in breast cancer. KLK8 is an independent indicator of prognosis in breast cancer patients \cite{michaelidou2015clinical}. Furthermore, t-SNE visualizations on the features from different modalities of all samples $\{\{f_i^m\}_{m=1}^M\}_{i=1}^N$ are performed to show the informativeness of the features that Multi-QuAD learns. Fig. \ref{fig:tsne-brca} and \ref{fig:tsne-rosmap} show the results on the BRCA and ROSMAP datasets. The features of all modalities enhanced by Multi-QuAD exhibit minimal intra-class distances and clear inter-class separation, which indicate that Multi-QuAD effectively retains informative features and thus achieves reliability.

\begin{table}[t]
\caption{Top 5 informative biomarkers that Multi-QuAD learned from different modalities on BRCA dataset.}
\begin{center}
\begin{tabular}{c|c}
\hline
Modality & Top 5 informative biomarkers\\ \hline
DNA & ZNF671, KRTAP3-1, ZCCHC8, AGR2, MYT1 \\ \hline
mRNA & KLK8, MIA, PTX3, SOX11, KRT6B \\ \hline
miRNA & mir-378, mir-190b, mir-934, mir-21, mir-23b \\
\hline
\end{tabular}
\end{center}
\label{tab_fea_learn}
\end{table}

\begin{table}[t]
\caption{Comparison of ACC, FLOPs per sample, and \#param of different methods on the FOOD101 dataset, where $*$ are the results under Gaussian noise with $\sigma=5$.}
\begin{center}
\begin{adjustbox}{width=0.95\linewidth}
\begin{tabular}{c|c|c|c|c}
\hline
Method & GCFANet & PDF & DynMM & Multi-QuAD\\ \hline
ACC (\%) & 92.9 & 93.3 & 89.3 & \textbf{94.5} \\ \hline
FLOPs (M) & 9.22 & 3.53 & 3.71 & 3.82 \\ \hline
\#param (M) & 9.13 & 3.46 & 3.67 & 3.77 \\ \hline
ACC$^*$ (\%) & 89.4 (-3.5) & 90.1 (-3.2) & 70.1 (-19.2) & \textbf{93.8} (-0.7) \\ \hline
FLOPs$^*$ (M) & 9.22 (+0) & 3.53 (+0) & 10.82 (+7.11) & 8.94 (+5.12) \\ \hline
\#param$^*$ (M) & 9.13 (+0) & 3.46 (+0) & 10.53 (+6.86) & 8.91 (+5.14) \\ \hline
\end{tabular}
\end{adjustbox}
\end{center}
\label{tab_compute_effe}
\end{table}

\subsection{Computation Effectiveness of Multi-QuAD} \label{sec:dis-ce}
Extensive experiments also demonstrate that Multi-QuAD achieves high computational efficiency while ensuring reliability. Table \ref{tab_compute_effe} compares the classification accuracy (ACC), FLOPs, and parameter amount (\#param) of Multi-QuAD with some SOTA reliable multimodal classification methods and the multimodal dynamic neural network \textbf{DynMM} on both clean and noisy datasets. Multi-QuAD achieves the most stable and promising results with only a 5.12M increase in FLOPs and a 5.14M increase in \#param (less than \textbf{DynMM} and still lower than \textbf{GCFANet} after the increase) under noise.

\section{Conclusions}
This work proposes a novel multi-level quality-adaptive dynamic multimodal network (Multi-QuAD) for reliable multimodal classification. Multi-QuAD adopts a reliable multi-level quality estimation method, \textbf{\textit{NFCE}}, based on noise-free prototypes and classifier-free design. Additionally, Multi-QuAD proposes two mechanisms, termed \textbf{\textit{GCND}} and \textbf{\textit{LGP}}, to achieve sample-adaptive network depth and parameters based on modality and feature-level quality, thereby significantly enhancing inference reliability. Extensive experiments on four datasets have demonstrated Multi-QuAD's significant superiority over other state-of-the-art methods in terms of classification performance and reliability. The adaptability of Multi-QuAD to diverse data quality, the effectiveness of its components (\textbf{\textit{NFCE}}, \textbf{\textit{GCND}}, and \textbf{\textit{LGP}}), and its high computational efficiency have also been verified and discussed.


Existing multimodal learning methods have not been sufficiently studied for their reliability in other machine learning tasks, such as regression and semantic segmentation. This represents a novel and worthwhile problem for in-depth exploration. Therefore, future research in this article will be further extended and validated on a more diverse range of machine learning tasks.

\bibliographystyle{IEEEtran}
\bibliography{reference}

\vfill

\end{document}